\documentclass[10pt,journal,compsoc]{IEEEtran}

\ifCLASSOPTIONcompsoc
  \usepackage[nocompress]{cite}
\else
  \usepackage{cite}
\fi

\ifCLASSINFOpdf
\else
\fi

\usepackage{times}
\usepackage{epsfig}
\usepackage{graphicx}
\usepackage{amsmath}
\usepackage{amssymb}

\usepackage{enumitem}
\usepackage{bm}
\usepackage{blindtext}
\usepackage{caption}

\usepackage{microtype}

\usepackage[pagebackref=true,breaklinks=true,letterpaper=true,colorlinks,bookmarks=false]{hyperref}

\begin{document}
\title{Adversarially Robust One-class \\ Novelty Detection}

\author{Shao-Yuan~Lo,~\IEEEmembership{Student~Member,~IEEE,}
        Poojan~Oza,~\IEEEmembership{Student~Member,~IEEE,}
        and~Vishal~M.~Patel,~\IEEEmembership{Senior~Member,~IEEE}% <-this % stops a space
\IEEEcompsocitemizethanks{\IEEEcompsocthanksitem S.-Y. Lo, P. Oza, and V. M. Patel are with the Department
of Electrical and Computer Engineering, Johns Hopkins University,
Baltimore, MD, 21218.\protect\\
E-mail: \{sylo, poza2, vpatel36\}@jhu.edu}}
%\IEEEcompsocthanksitem J. Doe and J. Doe are with Anonymous University.}% <-this % stops a space

\markboth{IEEE Transactions on Pattern Analysis and Machine Intelligence, 2022}%
{Lo \MakeLowercase{\textit{et al.}}: Adversarially Robust One-class Novelty Detection}

\IEEEtitleabstractindextext{%
\begin{abstract}
One-class novelty detectors are trained with examples of a particular class and are tasked with identifying whether a query example belongs to the same known class. Most recent advances adopt a deep auto-encoder style architecture to compute novelty scores for detecting novel class data. Deep networks have shown to be vulnerable to adversarial attacks, yet little focus is devoted to studying the adversarial robustness of deep novelty detectors. In this paper, we first show that existing novelty detectors are susceptible to adversarial examples. We further demonstrate that commonly-used defense approaches for classification tasks have limited effectiveness in one-class novelty detection. Hence, we need a defense specifically designed for novelty detection. To this end, we propose a defense strategy that manipulates the latent space of novelty detectors to improve the robustness against adversarial examples. The proposed method, referred to as Principal Latent Space (PrincipaLS), learns the incrementally-trained cascade principal components in the latent space to robustify novelty detectors. PrincipaLS can purify latent space against adversarial examples and constrain latent space to exclusively model the known class distribution. We conduct extensive experiments on eight attacks, five datasets and seven novelty detectors, showing that PrincipaLS consistently enhances the adversarial robustness of novelty detection models. Code is available at \url{https://github.com/shaoyuanlo/PrincipaLS}
\end{abstract}

}

\maketitle

\IEEEdisplaynontitleabstractindextext

\IEEEpeerreviewmaketitle

\ifCLASSOPTIONcompsoc
\IEEEraisesectionheading{\section{Introduction}\label{sec:introduction}}
\else
\section{Introduction}
\label{sec:introduction}
\fi
%\IEEEPARstart{T}{his} demo file is intended to serve as a ``starter file''
%for IEEE Computer Society journal papers produced under \LaTeX\ using
%IEEEtran.cls version 1.8b and later.

\IEEEPARstart{O}{ne-class} novelty detection refers to the problem of determining if a test data sample is normal (known class) or anomalous (novel class). In real-world applications, novel data is difficult to collect since they are often rare or unsafe. Hence, one-class novelty detection considers training data from only a single known class. Most recent advances in one-class novelty detection are based on the deep Auto-Encoder (AE) style architectures, such as Denoising Auto-Encoder (DAE) \cite{salehi2021arae,vincent2008extracting}, Variational Auto-Encoder (VAE) \cite{kingma2013auto}, Adversarial Auto-Encoder (AAE) \cite{makhzani2015adversarial, pidhorskyi2018generative}, Generative Adversarial Network (GAN) \cite{goodfellow2014generative,perera2019ocgan,sabokrou2018adversarially,salehi2020puzzle,zhangp}, etc. Given an AE that learns the distribution of the known class, normal data are expected to be reconstructed accurately, while anomalous data are not. The reconstruction error of the AE is then used as a score for a test example to perform novelty detection. Although deep novelty detection methods achieve impressive performance, their robustness against adversarial attacks \cite{goodfellow2015explaining,Szegedy2014Intriguing} lacks exploration.

\begin{figure}[!t]
	\centering
	\includegraphics[width=0.48\textwidth]{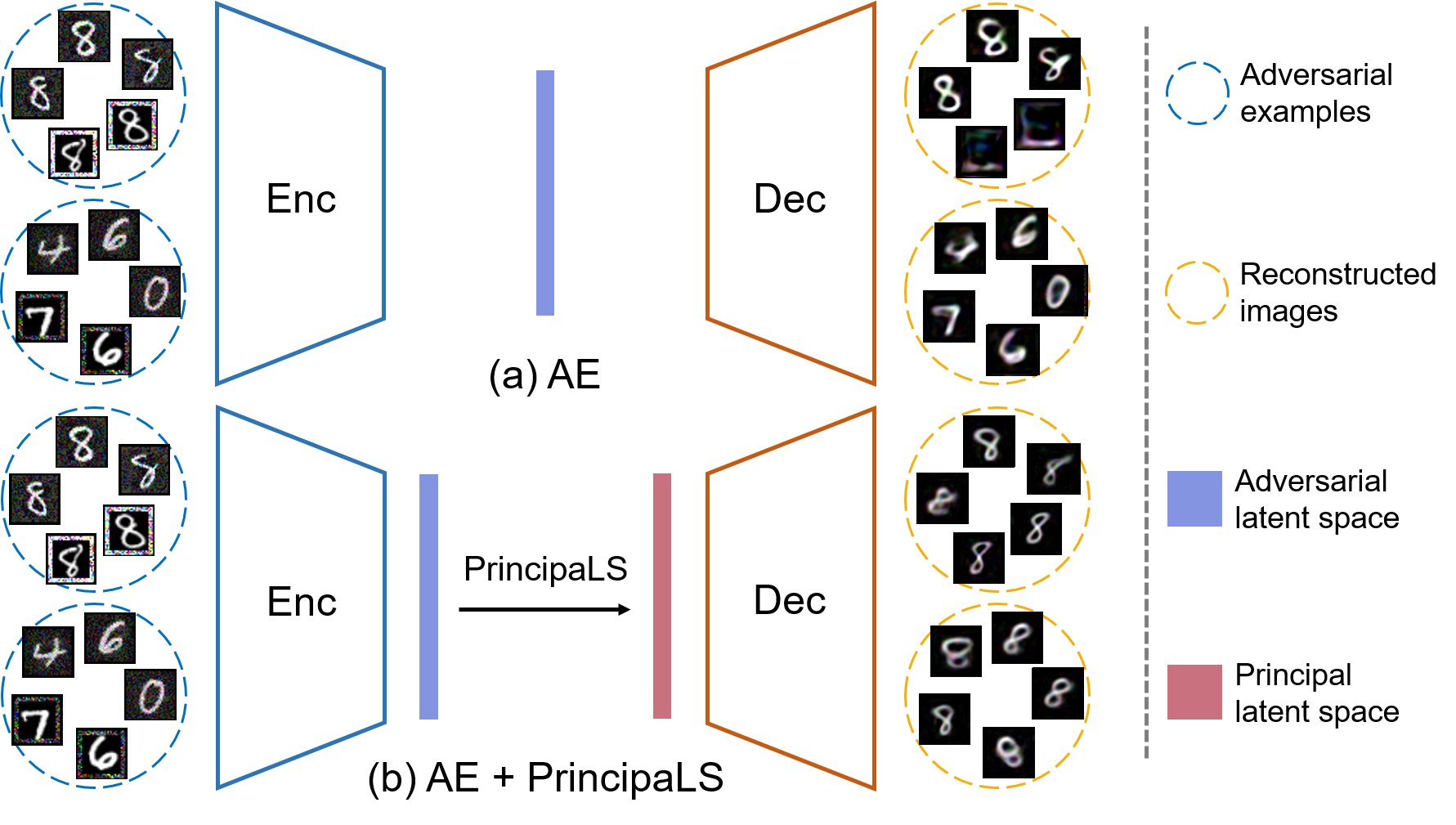}
	\caption{Overview of the proposed adversarially robust one-class novelty detection idea (PrincipaLS). The vanilla Auto-Encoder (AE) and AE+PrincipaLS are trained with the known class defined as digit 8. AE+PrincipaLS reconstructs every adversarial data into the known class (digit 8) and thus produces preferred reconstruction errors for novelty detection, even under attacks.}
	\label{fig:cover_pic}
\end{figure}

Adversarial examples pose serious security threats to deep networks as they can fool them with carefully crafted perturbations. Over the past few years, many adversarial attack and defense approaches have been proposed for tasks such as image classification \cite{guo2017countering,raff2019barrage,Xie_2019_CVPR,xu2017feature}, video recognition \cite{lo2021defending,wei2019sparse}, optical flow estimation \cite{ranjan2019attacking} and open-set recognition \cite{shao2020open}. However, adversarial attacks or defenses have not been thoroughly investigated in the context of one-class novelty detection. We first show that present novelty detectors are vulnerable to adversarial attacks. Subsequently, we demonstrate that many state-of-the-art defenses \cite{hendrycks2019selfsupervised,shi2021online,xie2020smooth,Xie_2019_CVPR} prove to be sub-optimal to properly defend novelty detectors against adversarial examples. This motivates us to design an effective defense strategy specifically for one-class novelty detection.

To this end, we propose to leverage task-specific knowledge to protect novelty detectors. These novelty detectors are only required to retain information about normal data, thereby resulting in poor reconstructions for anomalous data. This is favorable to the novelty detection problem. This can be achieved by constraining the latent space to make the features closer to a prior distribution \cite{perera2019ocgan,park2020learning}. Furthermore, it has been shown that adversarial perturbations can be removed in the feature space \cite{Xie_2019_CVPR}. Therefore, one can largely manipulate the latent space of novelty detectors to devoid them of feature corruption introduced by adversaries, while maintaining the performance on clean input data. This property is unique to the novelty detection task, as most deep learning applications (e.g., image classification) require a model containing sophisticated semantic information, and a large manipulation on the latent space may limit the model capability, resulting in performance degradation.

In this paper, we propose a defense strategy, referred to as Principal Latent Space (PrincipaLS), to defend novelty detectors against adversarial examples. Specifically, PrincipaLS learns the incrementally-trained \cite{ross2008incremental} cascade principal components in the latent space. This contains a cascade principal component analysis (PCA), which consists of a PCA operating on the vector dimension (i.e., channel) of a latent space \cite{van2017neural} and the other PCA operating on the spatial dimension. We name these two PCAs as \textit{Vector-PCA} and \textit{Spatial-PCA}, respectively. First, Vector-PCA uses a learned \textit{principal latent vector} to represent a latent space as the Vector-PCA space of a single-channel map. Since the principal latent vector is a pre-trained component that would not be affected by adversarial perturbations, most adversaries are removed at this step, and the remaining adversaries are enclosed within the small \textit{Vector-PCA space}. Subsequently, Spatial-PCA uses learned \textit{principal Vector-PCA maps} to represent the Vector-PCA space as the \textit{Spatial-PCA space} and expel the remaining adversaries. Finally, the corresponding cascade inverse PCA transforms the Spatial-PCA space back to the original dimensionality, resulting in the \textit{principal latent space}.

With PrincipaLS, the decoder could compute preferred reconstruction errors as novelty scores, even under adversarial attacks (see Fig.~\ref{fig:cover_pic}). Additionally, we incorporate adversarial training (AT) \cite{madry2018towards} with PrincipaLS to further exert PrincipaLS's ability in enhancing adversarial robustness. In contrast to typical defenses which often sacrifice their performance on clean data \cite{tsipras2018robustness,xie2020adversarial}, the proposed defense strategy does not hurt the performance but rather improves it. The PrincipaLS module can be attached to any AE-style architectures (VAE, GAN, etc), so it can be applied to a wide variety of the existing novelty detection approaches, such as \cite{kingma2013auto,makhzani2015adversarial,sabokrou2018adversarially,pidhorskyi2018generative,salehi2021arae} etc. Moreover, the PrincipaLS module is light-weight and computationally efficient.

We establish a solid evaluation benchmark for the problem of adversarially robust one-class novelty detection. We extensively evaluate PrincipaLS on \textbf{eight} adversarial attacks (ranging from digital to physically realizable attacks and from white-box to black-box attacks), \textbf{five} datasets (ranging from toy to realistic datasets and from image to video datasets) and \textbf{seven} different novelty detectors. We further compare PrincipaLS with commonly-used defense methods and show that it consistently enhances the adversarial robustness of novelty detectors by significant margins. To the best of our knowledge, this is one of the first adversarially robust novelty detection methods. We hope that the provided evaluation benchmark and comprehensive baseline results for this emerging problem will be useful to the vision and machine learning communities.

Our main contributions are summarized as follows:
\begin{itemize}
	\item We propose a novel adversarial defense method, PrincipaLS, based on task-specific knowledge to protect novelty detectors. To the best of our knowledge, this is one of the first adversarially robust novelty detection methods.
	\item We establish a solid evaluation benchmark for the problem of adversarially robust novelty detection.
	\item The proposed PrincipaLS consistently enhances the adversarial robustness of novelty detectors by wide margins. This holds true on multiple attacks, datasets and novelty detectors.
	\item We provide extensive analysis and discussion to study the proposed method and this emerging problem.
	\item We provide comprehensive baseline results for this emerging problem. These baselines broadly cover eight adversarial attacks, five datasets and seven different novelty detectors.
\end{itemize}

\section{Related work}  \label{sec2}
\noindent \textbf{One-class novelty detection.}
One-class novelty detection is of great interest to the computer vision community.  Earlier algorithms mainly rely on Support Vector Machines (SVM) formulation \cite{scholkopf1999support,tax2004support}. With the advent of deep learning, AE-based approaches are dominating this area and achieve state-of-the-art performance \cite{gong2019memorizing,park2020learning,perera2019ocgan,pidhorskyi2018generative,sabokrou2018adversarially,sakurada2014anomaly,salehi2020puzzle,salehi2021arae,xia2015learning,zhou2017anomaly}. ALOCC \cite{sabokrou2018adversarially} considers a DAE \cite{vincent2008extracting} as a generator and appends a discriminator to train the entire network by the GAN framework \cite{goodfellow2014generative}. GPND \cite{pidhorskyi2018generative} is based on AAE \cite{makhzani2015adversarial}, and it employs a discriminator to the latent space and the other discriminator to the output. OCGAN \cite{perera2019ocgan} includes two discriminators and a classifier to train a DAE by the GAN framework. ARAE \cite{salehi2021arae} crafts adversarial examples from the latent space to adversarially train a DAE. Puzzle-AE \cite{salehi2020puzzle} uses puzzle-solving as a pretext task to learn useful features, and it also incorporates adversarially robust training and the GAN training framework. Different from our work, ARAE and Puzzle-AE’s adversarial examples aim to pursue performance, and their adversarial robustness is not thoroughly evaluated (see Sec.~\ref{sec:arae}).

\noindent \textbf{Adversarial attacks.}
Szegedy et al. \cite{Szegedy2014Intriguing} showed that carefully crafted perturbations can fool deep networks. Goodfellow et al. \cite{goodfellow2015explaining} introduced the Fast Gradient Sign Method (FGSM), which leverages the sign of gradients to produce adversarial examples. Projected Gradient Descent (PGD) \cite{madry2018towards} extends FGSM from single iteration gradient descent to an iterative version. MI-FGSM \cite{dong2018boosting} generates more transferable adversarial attacks by a momentum mechanism. MultAdv \cite{lo2021multav} produces adversarial examples via the multiplicative operation instead of the additive operation. Physically realizable attacks, which can be implemented in the physical scenarios, is also developed. For example, Adversarial Framing (AF) \cite{zajac2019adversarial} adds perturbations on the border of an image, while the remaining pixels are unchanged.

\noindent \textbf{Adversarial defenses.}
Earlier approaches focus on detecting adversarial examples \cite{hendrycks2016early,jere2020principal,li2017adversarial}. However, it is well-known that detection is inherently weaker than defense in terms of resisting adversarial attacks. Although several defense approaches based on image transformation are proposed \cite{guo2017countering,xu2017feature,bhagoji2017dimensionality}, they fail to defend against white-box attacks \cite{carlini2017adversarial,obfuscated}. Recently, Adversarial Training (AT) has been considered one of the most effective defenses, especially in the white-box setting. Madry et al. \cite{madry2018towards} formulated AT in a min-max optimization framework (PGD-AT), and this has been widely used as a benchmark. Xie et al. \cite{Xie_2019_CVPR} includes the feature denoising block (FD) in networks to remove adversarial perturbations in the feature domain. SAT \cite{xie2020smooth} uses smooth approximations of ReLU activation to enhance PGD-AT. Hendrycks et al. \cite{hendrycks2019selfsupervised} added an auxiliary rotation prediction task \cite{gidaris2018unsupervised} to improve PGD-AT (RotNet-AT). SOAP \cite{shi2021online} takes self-supervised signals to purify adversarial examples during inference.

To the best of our knowledge, APAE \cite{goodge2020robustness} might be the only present defense designed for anomaly detection. It uses approximate projection and feature weighting to reduce adversarial effects. However, its robustness is not fully tested and only anomalous data are perturbed in its evaluation (see Sec.~\ref{sec:apae}). Instead, we provide a generic framework for evaluating the adversarial robustness of novelty detectors and our proposed defense method.

\section{Attacking novelty detection models} \label{sec3}

We consider several popular adversarial attacks \cite{dong2018boosting,goodfellow2015explaining,lo2021multav,madry2018towards,papernot2017practical,zajac2019adversarial} and modify their loss objectives to suit the novelty detection problem setup. Here, we take PGD \cite{madry2018towards} as an example to illustrate our attack formulation. The other gradient-based attacks can be extended by a similar formulation (see Sec.~\ref{sec:more_attacks} in the Supplementary).

Consider an AE-based target model with an encoder $Enc$ and a decoder $Dec$, and an input image $\mathbf{X}$ with the ground-truth label $y \in \{-1, 1\}$, where ``$1$" denotes the known class and ``$-1$" denotes the novel classes. We generate the adversarial example $\mathbf{X}_{adv}$ as follows:
\begin{equation}  \label{pgd_attack}
\mathbf{X}^{t+1} = Proj^{L_\infty}_{\mathbf{X}, \ \epsilon} \big\{ \mathbf{X}^{t} + \alpha \cdot sign(\bigtriangledown_{\mathbf{X}^t} \mathcal{L}(\hat{\mathbf{X}}^t, \mathbf{X}^t, y)) \big\},
\end{equation}
where, $\hat{\mathbf{X}}^t = Dec(Enc(\mathbf{X}^t))$, $\alpha>0$ denotes a step size, and $t \in [0,t_{max}-1]$ is the number of attacking iterations, $\mathbf{X} = \mathbf{X}^0$ and $\mathbf{X}_{adv} = \mathbf{X}^{t_{max}}$. $Proj^{L_\infty}_{\mathbf{X},\epsilon}\{\cdot\}$ projects its element into an $L_\infty$-norm bound with perturbation size $\epsilon$ such that $\parallel \mathbf{X}^{t+1} - \mathbf{X} \parallel_{\infty} \leq \epsilon$. $\mathcal{L}$ corresponds to the mean square error (MSE) loss defined as follows:
% MSE loss
\begin{equation}  \label{mse_loss}
\mathcal{L}(\hat{\mathbf{X}}^t, \mathbf{X}^t, y) = y \parallel \hat{\mathbf{X}}^t - \mathbf{X}^t \parallel_2.
\end{equation}
Given a test example, if it belongs to the known class, we maximize its reconstruction error (i.e., novelty score) by gradient ascent; while if it belongs to novel classes, we minimize its reconstruction error by gradient descent. We use this formulation to generate adversarial examples for doing AT as well. During AT, since we can only access the training data of the known class, the label $y$ is always $1$ in Eq.~\eqref{pgd_attack} and Eq.~\eqref{mse_loss}.

Present novelty detection methods are vulnerable to this attack (see Sec.~\ref{sec:robustness}); that is, normal data would be misclassified into novel classes, and anomalous data would be misclassified into the known class. Moreover, this attacking strategy is much stronger than the attacks introduced by \cite{salehi2021arae}, which perturbs only normal data, and by \cite{goodge2020robustness}, which perturbs only anomalous data. Because the proposed attack is stronger, our AT for defense is much more effective accordingly. A detailed comparison of the attacking strategies is discussed in Sec.~\ref{sec:arae} and Sec.~\ref{sec:apae}. The proposed strong attack establishes a solid evaluation benchmark for the problem of adversarially robust one-class novelty detection.
% Therefore, we can evaluate the adversarial robustness of novelty detection models properly.

\section{Adversarially robust novelty detection} \label{sec:method}
The proposed defense strategy exploits the task-specific knowledge of one-class novelty detection. Specifically, we leverage the fact that a novelty detector's latent space can be manipulated to a larger extent as long as it retains the known class information. This property is especially useful to remove more adversarial perturbations in the latent space. Therefore, we propose to train a novelty detector by manipulating its latent space such that it can improve adversarial robustness while maintaining the performance on clean data. Note that these characteristics are specific to the novelty detection problem. The majority of visual recognition problems, such as image classification, require a model retaining multiple category information. Hence, a large manipulation on the latent space may hinder the model capability and thus degrade the performance. In the following subsections, we first briefly review PCA to define the notations used in this paper, then discuss the proposed PrincipaLS in detail.
%The proposed method PrincipaLS transforms the latent space to the principal latent space. PrincipaLS is based on an incrementally-trained cascade PCA, which consists of Vector-PCA and Spatial-PCA we call. 

\subsection{Preliminary}
PCA computes the principal components of a collection of data and uses them to conduct a change of basis on the data through a linear transformation. Consider a data matrix $\mathbf{X} \in \mathbb{R}^{n \times d}$, its mean $\bm{\mu} \in \mathbb{R}^{1 \times d}$ and its covariance $\mathbf{C} = (\mathbf{X}-\bm{\mu})^\top (\mathbf{X}-\bm{\mu})$. $\mathbf{C}$ can be written as $\mathbf{C} = \mathbf{U} \bm{\Lambda} \mathbf{V}^\top$ via Singular Vector Decomposition (SVD), where $\mathbf{U} \in \mathbb{R}^{d \times d}$ is an orthogonal matrix containing the principal components of $\mathbf{X}$. Here, we define a mapping $h$ which computes the mean vector and the first $k$ principal components of the given $\mathbf{X}$:
% h(X)
\begin{equation}
\label{hhh}
h(\mathbf{X}, k): \mathbf{X} \to \{\bm{\mu}, \tilde{\mathbf{U}}\},
\end{equation}
where $\tilde{\mathbf{U}} \in \mathbb{R}^{d \times k}$ keeps only the first $k$ columns of $\mathbf{U}$. Now we define the forward and the inverse PCA transformation as a pair of mapping $(f: \mathbb{R}^{n \times d} \to \mathbb{R}^{n \times k}$, $g: \mathbb{R}^{n \times k} \to \mathbb{R}^{n \times d})$; $f$ performs the forward PCA: 
% f(x)
\begin{equation}
\label{fff}
f(\mathbf{X}; \bm{\mu}, \tilde{\mathbf{U}}) = (\mathbf{X}-\bm{\mu}) \tilde{\mathbf{U}},
\end{equation}
and $g$ performs the inverse PCA:
% g(x)
\begin{equation}
\label{ggg}
g(\mathbf{X}_{PCA}; \bm{\mu}, \tilde{\mathbf{U}}) = \mathbf{X}_{PCA} \tilde{\mathbf{U}}^\top + \bm{\mu},
\end{equation}
where $\mathbf{X}_{PCA} = f(\mathbf{X}; \bm{\mu}, \tilde{\mathbf{U}})$. Finally, we can write the PCA reconstruction of $\mathbf{X}$ as $\hat{\mathbf{X}} = g(f(\mathbf{X}; \bm{\mu}, \tilde{\mathbf{U}}); \bm{\mu}, \tilde{\mathbf{U}})$.

\subsection{Principal Latent Space (PrincipaLS)}
The proposed PrincipaLS contains two major components: (1) Vector-PCA and (2) Spatial-PCA. In Vector-PCA, we perform $(h, f, g)$ on the vector dimension as $(h_V, f_V, g_V)$, and in Spatial-PCA, we perform $(h, f, g)$ on the spatial dimension as $(h_S, f_S, g_S)$. Let $Enc$ be the encoder and $Dec$ be the decoder of a novelty detection model. Let us denote an adversarial image as $\mathbf{X}_{adv}$, we have its latent space $\mathbf{Z}_{adv} = Enc(\mathbf{X}_{adv}) \in \mathbb{R}^{s \times v}$, where $s = h \times w$ is the spatial dimensionality obtained by the product of height and width, and $v$ is the vector dimensionality (i.e., the number of channels). Under adversarial attacks, $\mathbf{Z}_{adv}$ would be corrupted by adversarial perturbations such that the decoder cannot compute reconstruction errors favorable to novelty detection. We define the proposed PrincipaLS as a transformation $PrincipaLS: \mathbf{Z}_{adv} \to \mathbf{Z}_{PrincipaLS}$, which removes adversaries from $\mathbf{Z}_{adv}$, where $\mathbf{Z}_{PrincipaLS}$ is referred to as principal latent space. $PrincipaLS$ is implemented by our incrementally-trained cascade PCA. In the beginning, a sigmoid function replaces the encoder's last activation function to bound $\mathbf{Z}_{adv}$ values between 0 and 1. The following procedures are described below.

\begin{figure}[!t]
	\centering
	\includegraphics[width=0.48\textwidth]{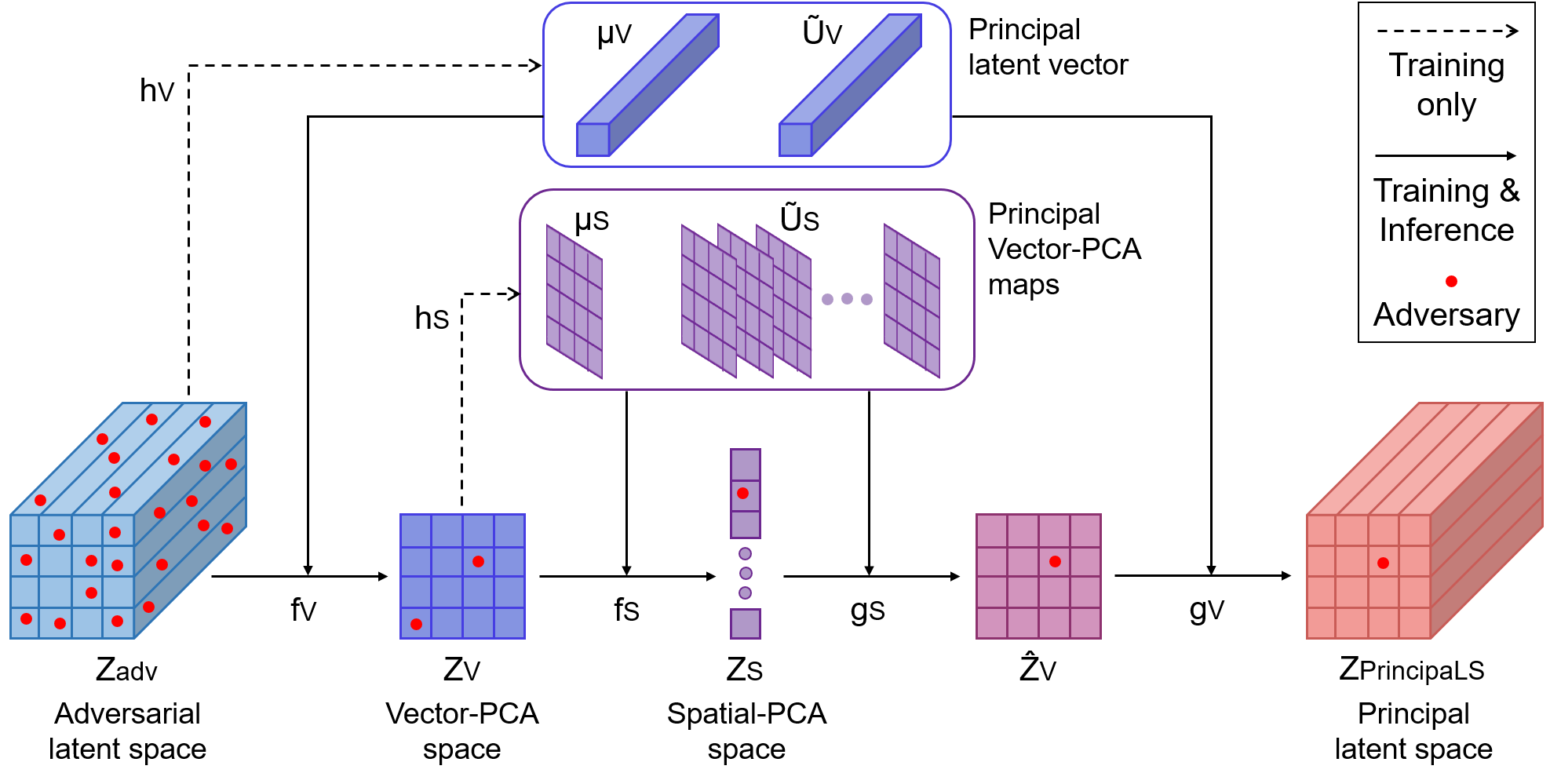}
	\caption{Overview of the proposed PrincipaLS. $f_V$: forward Vector-PCA, $f_S$: forward Spatial-PCA, $g_S$: inverse Spatial-PCA, $g_V$: inverse Vector-PCA, $h_V$ and $h_S$ are the mappings for computing principal components.}
	\label{fig:big_pic}
\end{figure}

First, Vector-PCA computes the mean latent vector and the principal latent vector of $\mathbf{Z}_{adv}$:
% hv(z)
\begin{equation}
\label{hv}
\{\bm{\mu}_V, \tilde{\mathbf{U}}_V\} = h_V(\mathbf{Z}_{adv}, k_V=1),
\end{equation}
where, we always set $k_V$ to 1, so $\tilde{\mathbf{U}}_V$ is the first principal latent vector of $\mathbf{Z}_{adv}$. Second, Vector-PCA transforms $\mathbf{Z}_{adv}$ to its Vector-PCA space $\mathbf{Z}_V \in \mathbb{R}^{s \times 1}$:
% fv(z)
\begin{equation}
\label{fv}
\mathbf{Z}_V = f_V(\mathbf{Z}_{adv}; \bm{\mu}_V, \tilde{\mathbf{U}}_V).
\end{equation}
Next, Spatial-PCA computes the mean Vector-PCA map\footnote{We use the word ``map" to indicate they are on the spatial dimension.} and the principal Vector-PCA maps of $\mathbf{Z}_V$:
% hs(z)
\begin{equation}
\label{hs}
\{\bm{\mu}_S, \tilde{\mathbf{U}}_S\} = h_S(\mathbf{Z}_V^\top, k_S),
\end{equation}
where, $k_S$ is a hyperparameter. Then, Spatial-PCA transforms $\mathbf{Z}_V$ to its Spatial-PCA space $\mathbf{Z}_S \in \mathbb{R}^{k_S \times 1}$:
% fs(z)
\begin{equation}
\label{fs}
\mathbf{Z}_S^\top = f_S(\mathbf{Z}_V^\top; \bm{\mu}_S, \tilde{\mathbf{U}}_S).
\end{equation}
Finally, the inverse Spatial-PCA and the inverse Vector-PCA transform $\mathbf{Z}_S$ back to its original dimensionality:
% gs(z)
\begin{equation}
\label{gs}
\hat{\mathbf{Z}}_V^\top = g_S(\mathbf{Z}_S^\top; \bm{\mu}_S, \tilde{\mathbf{U}}_S),
\end{equation}
% gv(z)
\begin{equation}
\label{gv}
\mathbf{Z}_{PrincipaLS} = g_V(\hat{\mathbf{Z}}_V; \bm{\mu}_V, \tilde{\mathbf{U}}_V),
\end{equation}
where, $\hat{\mathbf{Z}}_V$ is the Spatial-PCA reconstruction of $\mathbf{Z}_V$, and $\mathbf{Z}_{PrincipaLS}$ is the resulting principal latent space. Fig.~\ref{fig:big_pic} gives an overview of this procedure. The decoder then uses $\mathbf{Z}_{PrincipaLS}$ to reconstruct the input adversarial example as $\hat{\mathbf{X}}_{adv} = Dec(\mathbf{Z}_{PrincipaLS})$ for computing the novelty score.

\subsection{Incremental training}
The \textit{principal latent components} $\{\bm{\mu}_V, \tilde{\mathbf{U}}_V, \bm{\mu}_S, \tilde{\mathbf{U}}_S\}$ are incrementally-trained along with the network weights by exponential moving average (EMA) during training, so we call this process  incrementally-trained cascade PCA. Specifically, at training iteration $t$, these components are updated with following equations:
% Train Vector-PCA
\begin{equation}
\label{train_V}
\{\bm{\mu}_V^t, \tilde{\mathbf{U}}_V^t\} = (1-\eta_V) \{\bm{\mu}_V^{t-1}, \tilde{\mathbf{U}}_V^{t-1}\} + \eta_V \cdot h_V(\mathbf{Z}_{adv}^t),
\end{equation}
% Train Spatial-PCA
\begin{equation}
\label{train_S}
\{\bm{\mu}_S^t, \tilde{\mathbf{U}}_S^t\} = (1-\eta_S) \{\bm{\mu}_S^{t-1}, \tilde{\mathbf{U}}_S^{t-1}\} + \eta_S \cdot h_S(\mathbf{Z}_V^{t \top}),
\end{equation}
where $\eta_V$ and $\eta_S$ are the EMA learning rates.

Consider the model weights are trained by the mini-batch gradient descent with a batch size $b$, the latent dimensionality is shaped to $\mathbf{Z}_{adv} \in \mathbb{R}^{bs \times v}$, the resulting $\mathbf{Z}_V \in \mathbb{R}^{bs \times 1}$ is reshaped to $\mathbf{Z}_V \in \mathbb{R}^{s \times b}$ after the Vector-PCA $f_V$, and $\hat{\mathbf{Z}}_V \in \mathbb{R}^{s \times b}$ is reshaped back to $\hat{\mathbf{Z}}_V \in \mathbb{R}^{bs \times 1}$ after the inverse Spatial-PCA $g_S$. Hence, in a mini-batch, both $h_V$ and $h_S$ have $b$ times more data points to acquire better principal latent components at each training iteration. At iteration $t$, $(f_V, g_V)$ performs with the components $\{\bm{\mu}_V^t, \tilde{\mathbf{U}}_V^t\}$, and $(f_S, g_S)$ performs with the components $\{\bm{\mu}_S^t, \tilde{\mathbf{U}}_S^t\}$. When the training process ends, the well-trained components are denoted as $\{\bm{\mu}_V^*, \tilde{\mathbf{U}}_V^*, \bm{\mu}_S^*, \tilde{\mathbf{U}}_S^*\}$. During infernce, $(f_V, g_V)$ performs with $\{\bm{\mu}_V^*, \tilde{\mathbf{U}}_V^*\}$, and $(f_S, g_S)$ performs with $\{\bm{\mu}_S^*, \tilde{\mathbf{U}}_S^*\}$, while $h_V$ and $h_S$ do not operate (see Fig.~\ref{fig:big_pic}). The entire process is differentiable during inference and thus does not cause obfuscated gradients \cite{obfuscated}.

This incremental training helps make sure that the cascade PCA is aware of the network weight updates at each training step, and vice versa \cite{van2017neural}. Therefore, when one is updated, the other one would be updated accordingly. The incremental training encourages mutual learning between the principle latent components and the network weights. The entire model and thus can be well-trained end-to-end.

\subsection{Defense mechanism}
% (in the sense of the inverse Vector-PCA)
We further elaborate on how the proposed PrincipaLS defends against adversarial attacks. Given an adversarial example $\mathbf{X}_{adv}$, its latent space $\mathbf{Z}_{adv}$ is adversarially perturbed. After Vector-PCA, each latent vector of $\mathbf{Z}_{adv}$ is represented by a scaling factor of the learned principal latent vector $\tilde{\mathbf{U}}_V^*$ (with a bias term $\bm{\mu}_V^*$). The Vector-PCA space $\mathbf{Z}_V$ stores these scaling factors on a single-channel map (i.e., on the spatial domain only). Since all the principal latent components are pre-trained parameters, they would not be affected by adversarial perturbations. Replacing the perturbed latent vectors by $\tilde{\mathbf{U}}_V^*$ removes the majority of the adversaries. The only place where the remaining adversaries can appear is the scaling factors of $\tilde{\mathbf{U}}_V^*$ on the single-channel map. In other words, these adversaries are enclosed within a small subspace, making them easier to expel.

Subsequently, Spatial-PCA reconstructs this small subspace by a set of principal Vector-PCA maps $\tilde{\mathbf{U}}_S^*$ (with a bias term $\bm{\mu}_S^*$). Since $\tilde{\mathbf{U}}_S^*$ and $\bm{\mu}_S^*$ are adversary-free, the remaining adversaries are further removed. From another perspective, this step can be viewed as PCA-based denoising performing in the spatial domain of features. With the robust principal latent space $\mathbf{Z}_{PrincipaLS}$, the decoder can obtain a preferred reconstruction error for novelty detection, even in the presence of an adversarial example. Additionally, we perform AT \cite{madry2018towards} to train the model, further improving the robustness.

%On the other hand, $PrincipaLS$ reduces the dimensionality of the latent space and thus restricts the generalization of its reconstruction ability. This makes novelty detectors exclusively model the distribution of the known class. With such restriction, although the reconstruction errors of both normal and anomalous data increase, the error gap between normal data and anomalous data is enlarged. This benefits novelty detection even under attacks, and performance can improve simultaneously. Details are shown in Sec. \ref{ablation}.

\section{Experiments}
We evaluate PrincipaLS on eight adversarial attacks, five datasets and seven existing novelty detection methods. We further compare PrincipaLS with state-of-the-art defense approaches. Extensive analysis and discussion are also presented.

% Main results 1
\renewcommand{\arraystretch}{0.5}
\setlength{\tabcolsep}{6pt}
\begin{table*}[htp!]
	\begin{center}
		\caption{The mAUROC of models under various adversarial attacks.}
		\label{table:main_results_1}
		\begin{tabular}{r | r | c | ccccc | c | c}
			\hline \noalign{\smallskip} \noalign{\smallskip}
			Dataset & Defense & Clean & FGSM \cite{goodfellow2015explaining} & PGD \cite{madry2018towards} & MI-FGSM \cite{dong2018boosting} & MultAdv \cite{lo2021multav} & AF \cite{zajac2019adversarial} & Black-box \cite{papernot2017practical} & Average \\
			\noalign{\smallskip} \hline \noalign{\smallskip}
			& No Defense & 0.964 & 0.350 & 0.051 & 0.022 & 0.170 & 0.014 & 0.790 & 0.337 \\
			\noalign{\smallskip} \cline{2-10} \noalign{\smallskip}
			& PGD-AT \cite{madry2018towards} & 0.961 & 0.604 & 0.357 & 0.369 & 0.444 & 0.155 & 0.691 & 0.512 \\
			& FD \cite{Xie_2019_CVPR} & 0.963 & 0.612 & 0.366 & 0.379 & 0.453 & 0.142 & 0.700 & 0.516 \\
			MNIST & SAT \cite{xie2020smooth} & 0.947 & 0.527 & 0.295 & 0.306 & 0.370 & 0.142 & 0.652 & 0.463 \\
			\cite{lecun2010mnist} & RotNet-AT \cite{hendrycks2019selfsupervised} & 0.967 & 0.598 & 0.333 & 0.333 & 0.424 & 0.101 & 0.695 & 0.493 \\
			& SOAP \cite{shi2021online} & 0.940 & 0.686 & 0.504 & 0.506 & 0.433 & 0.088 & 0.863 & 0.574 \\
			& APAE \cite{goodge2020robustness} & 0.925 & 0.428 & 0.104 & 0.105 & 0.251 & 0.022 & 0.730 & 0.366 \\
			& PrincipaLS (ours) & \textbf{0.973} & \textbf{0.812} & \textbf{0.706} & \textbf{0.707} & \textbf{0.725} & \textbf{0.636} & \textbf{0.866} & \textbf{0.775} \\
			\noalign{\smallskip} \hline \noalign{\smallskip}
			& No Defense & 0.892 & 0.469 & 0.088 & 0.047 & 0.148 & 0.112 & 0.562 & 0.331 \\
			\noalign{\smallskip} \cline{2-10} \noalign{\smallskip}
			& PGD-AT \cite{madry2018towards} & 0.890 & 0.518 & 0.368 & 0.348 & 0.327 & 0.253 & 0.540 & 0.463 \\
			& FD \cite{Xie_2019_CVPR} & 0.886 & 0.524 & 0.379 & 0.359 & 0.335 & 0.252 & 0.535 & 0.467 \\
			F-MNIST & SAT \cite{xie2020smooth} & 0.878 & 0.444 & 0.306 & 0.285 & 0.273 & 0.231 & 0.492 & 0.416 \\
			\cite{xiao2017fashion} & RotNet-AT \cite{hendrycks2019selfsupervised} &  0.891 & 0.527 & 0.375 & 0.351 & 0.312 & 0.240 & 0.541 & 0.462 \\
			& SOAP \cite{shi2021online} & 0.876 & 0.639 & 0.475 & 0.475 & 0.327 & 0.274 & 0.611 & 0.525 \\
			& APAE \cite{goodge2020robustness} & 0.861 & 0.510 & 0.174 & 0.174 & 0.220 & 0.135 & 0.513 & 0.370 \\
			& PrincipaLS (ours) & \textbf{0.909} & \textbf{0.687} & \textbf{0.613} & \textbf{0.599} & \textbf{0.590} & \textbf{0.605} & \textbf{0.711} & \textbf{0.673} \\
			\noalign{\smallskip} \hline \noalign{\smallskip}
			& No Defense & 0.550 & 0.186 & 0.034 & 0.018 & 0.025 & 0.035 & 0.227 & 0.154 \\
			\noalign{\smallskip} \cline{2-10} \noalign{\smallskip}
			& PGD-AT \cite{madry2018towards} & 0.546 & 0.236 & 0.145 & 0.139 & 0.107 & 0.096 & 0.223 & 0.213 \\
			& FD \cite{Xie_2019_CVPR} & 0.546 & 0.237 & 0.147 & 0.141 & 0.109 & 0.103 & 0.222 & 0.215 \\
			CIFAR-10 & SAT \cite{xie2020smooth} & 0.537 & 0.223 & 0.141 & 0.135 & 0.101 & 0.079 & 0.219 & 0.205 \\
			\cite{krizhevsky2009learning} & RotNet-AT \cite{hendrycks2019selfsupervised} &  0.547 & 0.236 & 0.139 & 0.107 & 0.075 & 0.092 & 0.224 & 0.203 \\
			& SOAP \cite{shi2021online} & 0.546 & 0.270 & 0.131 & 0.141 & 0.096 & 0.070 & 0.231 & 0.211 \\
			& APAE \cite{goodge2020robustness} & 0.552 & 0.259 & 0.097 & 0.097 & 0.077 & 0.112 & 0.255 & 0.207 \\
			& PrincipaLS (ours) & \textbf{0.577} & \textbf{0.320} & \textbf{0.246} & \textbf{0.243} & \textbf{0.202} & \textbf{0.244} & \textbf{0.333} & \textbf{0.309} \\
			\noalign{\smallskip} \hline \noalign{\smallskip}
			& No Defense & 0.667 & 0.111 & 0.032 & 0.022 & 0.034 & 0.061 & \textbf{0.595} & 0.217 \\
			\noalign{\smallskip} \cline{2-10} \noalign{\smallskip}
			& PGD-AT \cite{madry2018towards} & 0.655 & 0.123 & 0.053 & 0.040 & 0.054 & 0.062 & 0.569 & 0.222 \\
			& FD \cite{Xie_2019_CVPR} & 0.658 & 0.145 & 0.061 & 0.050 & 0.061 & 0.066 & 0.572 & 0.230 \\
			MVTec-AD & SAT \cite{xie2020smooth} & 0.636 & 0.083 & 0.029 & 0.024 & 0.035 & 0.044 & 0.553 & 0.201 \\
			\cite{bergmann2019mvtec} & RotNet-AT \cite{hendrycks2019selfsupervised} & \textbf{0.677} & 0.123 & 0.050 & 0.038 & 0.049 & 0.059 & 0.586 & 0.226 \\
			& SOAP \cite{shi2021online} & 0.540 & 0.167 & 0.092 & 0.056 & 0.095 & \textbf{0.456} & 0.582 & 0.284 \\
			& APAE \cite{goodge2020robustness} & 0.621 & 0.142 & 0.058 & 0.044 & 0.058 & 0.120 & 0.553 & 0.228 \\
			& PrincipaLS (ours) & 0.638 & \textbf{0.334} & \textbf{0.243} & \textbf{0.238} & \textbf{0.197} & 0.164 & 0.542 & \textbf{0.337} \\
			\noalign{\smallskip} \hline \noalign{\smallskip}
			& No Defense & 0.523 & 0.204 & 0.034 & 0.038 & 0.006 & 0.000 & 0.220 & 0.146 \\
			\noalign{\smallskip} \cline{2-10} \noalign{\smallskip}
			& PGD-AT \cite{madry2018towards} & 0.527 & 0.217 & 0.168 & 0.154 & 0.100 & 0.000 & 0.221 & 0.198 \\
			& FD \cite{Xie_2019_CVPR} & 0.528 & 0.226 & 0.189 & 0.181 & 0.132 & 0.002 & 0.229 & 0.212 \\
			SHTech & SAT \cite{xie2020smooth} & \textbf{0.529} & 0.184 & 0.110 & 0.092 & 0.040 & 0.000 & 0.199 & 0.165 \\
			\cite{liu2018future} & RotNet-AT \cite{hendrycks2019selfsupervised} &  0.516 & 0.220 & 0.163 & 0.158 & 0.113 & 0.000 & 0.229 & 0.200 \\
			& SOAP \cite{shi2021online} & 0.432 & 0.024 & 0.002 & 0.000 & 0.002 & \textbf{0.181} & 0.202 & 0.120 \\
			& APAE \cite{goodge2020robustness} & 0.510 & 0.215 & 0.048 & 0.050 & 0.011 & 0.000 & 0.207 & 0.149 \\
			& PrincipaLS (ours) & 0.498 & \textbf{0.274} & \textbf{0.223} & \textbf{0.217} & \textbf{0.175} & 0.051 & \textbf{0.308} & \textbf{0.249} \\
			\noalign{\smallskip} \hline
		\end{tabular}
	\end{center}
\end{table*}

\subsection{Experimental setup}  \label{sec51}
\noindent \textbf{Datasets.}
We use five datasets for evaluation: MNIST \cite{lecun2010mnist}, Fashion-MNIST (F-MNIST) \cite{xiao2017fashion}, CIFAR-10 \cite{krizhevsky2009learning}, MVTec-AD \cite{bergmann2019mvtec} and ShanghaiTech (SHTech) \cite{liu2018future}. MNIST consists of grayscale handwritten digits from 0 to 9. It contains 60,000 training data and 10,000 test data. F-MNIST is composed of grayscale images from 10 fashion product categories. It comprises of 60,000 training data and 10,000 test data. CIFAR-10 consists of color images from 10 different classes. There are 50,000 training and 10,000 test images in this dataset. MVTec-AD is an anomaly detection dataset that consists of color images from 15 objects and textures categories. Each category contains normal and anomalous images with different types of defects. There are 3,629 training and 1,725 (467 normal and 1258 anomalous) test images in this dataset. SHTech is a video anomaly detection dataset that consists of videos from 13 scenes. It contains 274,515 training and 40,791 (23,465 normal and 17,326 anomalous) test frames. It is the largest dataset among existing anomaly detection benchmarks. In our experiments, we resize all the datasets to 32 $\times$ 32 during both training and testing.

\noindent \textbf{Evaluation protocol.}
For the MNIST, F-MNIST and CIFAR-10 datasets, which are originally created for image classification, we simulate a one-class novelty detection scenario by the following protocol. Given a dataset, each class is defined as the known class at a time, and a model is trained with the training data of this known class. During inference, the test data of the known class are considered normal, and the test data of the other classes (i.e., novel classes) are considered anomalous. We select the anomalous data from each novel class equally to constitute half of the test set, where the anomalous data within a novel class are selected randomly. Hence, our test set contains 50\% anomalous data, where each novel class accounts for the same proportion. The area under the Receiver Operating Characteristic curve (AUROC) value is used as the evaluation metric, where the ROC curve is obtained by varying the threshold of the novelty score. For each dataset, we report the mean AUROC (mAUROC) across its 10 classes.

For the MVTec-AD dataset, we conduct experiments on all the 15 categories and report mAUROC across these 15 categories. Similarly, for each category, we sample the anomalous data from each defect type equally to constitute half of the test set such that the test set contains 50\% anomalous data. For the SHTech dataset, we directly use its default test set as its normal-to-anomalous ratio is more balanced. Following \cite{mahadevan2010anomaly,sabokrou2018adversarially}, we report frame-level AUROC.

\noindent \textbf{Baseline defenses.}
To the best of our knowledge, APAE \cite{goodge2020robustness} might be the only present defense designed for anomaly detection. In addition to APAE, we implement five commonly-used defenses, which are originally designed for classification tasks, in the context of novelty detection. They are PGD-AT \cite{madry2018towards}, FD \cite{Xie_2019_CVPR}, SAT \cite{xie2020smooth}, RotNet-AT \cite{hendrycks2019selfsupervised} and SOAP \cite{shi2021online}, where FD, SAT and RotNet-AT incorporate PGD-AT. We use Gaussian non-local means \cite{buades2005non} for FD, Swish \cite{hendrycks2016gaussian} for SAT, and RotNet \cite{gidaris2018unsupervised} for SOAP. These are their well-performing versions.

\noindent \textbf{Benchmark novelty detectors.}
We apply PrincipaLS to seven novelty detection methods, including a vanilla AE, VAE \cite{kingma2013auto}, AAE \cite{makhzani2015adversarial}, ALOCC \cite{sabokrou2018adversarially}, GPND \cite{pidhorskyi2018generative}, ARAE \cite{salehi2021arae} and Puzzle-AE \cite{salehi2020puzzle}, where the vanilla AE is the default novelty detector if not otherwise specified. PrincipaLS is added after the last layer of the novelty detection models' encoder.

In order to evenly evaluate the adversarial robustness of these approaches, we unify their AE backbones into the following archirecture. The encoder consists of four 3 $\times$ 3 convolutional layers, where each of the first three layers are followed by a 2 $\times$ 2 max-pooling with stride 2. We use a base channel size of 64, and increase the number of channels by a factor of 2. The decoder mirrors the encoder but replaces every max-pooling by a bilinear interpolation with a factor of 2. All the convolutional layers are followed by a batch normalization layer \cite{ioffe2015batch} and ReLU. 

%%%%% For ALOCC, we take the reconstruction errors from its network $\mathcal{R}$ as the novelty score instead of the output of the network $\mathcal{D}$ as used in \cite{sabokrou2018adversarially}, as this achieves much better performance, giveing a stronger ground.

\noindent \textbf{Attack setting.}
We test adversarial robustness against five white-box attacks, inclduing FGSM \cite{goodfellow2015explaining}, PGD \cite{madry2018towards}, MI-FGSM \cite{dong2018boosting}, MultAdv \cite{lo2021multav} and AF \cite{zajac2019adversarial}, where PGD is the default attack if not otherwise specified. A black-box attack and two adaptive attacks \cite{papernot2017practical,tramer2020adaptive} are also considered. All the attacks are implemented based on the formulation in Sec.~\ref{sec3}.

For FGSM, PGD and MI-FGSM, we set $\epsilon$ to $25/255$ for MNIST, $16/255$ for F-MNIST, $8/255$ for CIFAR-10, $2/255$ for MVTec-AD, and $8/255$ for SHTech. For MultAdv, we set $\epsilon_m$ to $1.25$ for MNIST, $1.16$ for F-MNIST, $1.08$ for CIFAR-10, $1.02$ for MVTec-AD, and $1.08$ for SHTech. For AF, we set $\epsilon$ to $160/255$, $120/255$, $80/255$, $20/255$ and $80/255$ for MNIST, F-MNIST, CIFAR-10, MVTec-AD and SHTech, respectively. The framing width $w_{AF}$ is set to $1$. The number of attack iterations $t_{max}$ is set to $1$ for FGSM and $5$ for the other attacks. All the defenses that incorporate PGD-AT (i.e., PGD-AT, FD, SAT, RotNet-AT and our PrincipaLS) use the PGD setting described here for doing AT.

\noindent \textbf{Implementation details.}
We implement experiments by PyTorch \cite{paszke2019pytorch}. All the models are trained by Adam optimizer \cite{kingma2014adam} with initial learning rate $5e^{-5}$ and weight decay $1e^{-4}$ for 50 epochs (except that 10 epochs for SHTech), where the learning rate is decreased by a factor of 10 at the 20th and 40th epochs. The batch size is 128. For PrincipaLS, we set $k_V$ to $1$, $k_S$ to $8$, initial $\eta_V$ to $0.1$ and initial $\eta_S$ to $0.001$, where $\eta_V$ and $\eta_S$ are also decreased by a factor of 10 at the 20th and 40th epochs.

% Main results 2
\setlength{\tabcolsep}{6pt}
\begin{table*}[htp!]
	\begin{center}
		\caption{The mAUROC of models under PGD attack. Various novelty detectors are used.}
		\label{table:main_results_2}
		\begin{tabular}{r | r | c | ccccccc}
			\hline \noalign{\smallskip} \noalign{\smallskip}
			Dataset & Defense & Test type & AE & VAE \cite{kingma2013auto} & AAE \cite{makhzani2015adversarial} & ALOCC \cite{sabokrou2018adversarially} & GPND \cite{pidhorskyi2018generative} & ARAE \cite{salehi2021arae} & Puzzle-AE \cite{salehi2020puzzle} \\
			\noalign{\smallskip} \hline \noalign{\smallskip}
			& No Defense & Clean & 0.964 & 0.979 & 0.973 & 0.961 & 0.946 & 0.965 & 0.967 \\
			& No Defense & PGD & 0.051 & 0.087 & 0.056 & 0.141 & 0.128 & 0.133 & 0.295 \\
			\noalign{\smallskip} \cline{2-10} \noalign{\smallskip}
			& PGD-AT \cite{madry2018towards} & & 0.357 & 0.521 & 0.427 & 0.312 & 0.582 & 0.341 & 0.319 \\
			MNIST & FD \cite{Xie_2019_CVPR} & & 0.366 & 0.525 & 0.419 & 0.319 & 0.551 & 0.350 & 0.322 \\
			\cite{lecun2010mnist} & SAT \cite{xie2020smooth} & & 0.295 & 0.485 & 0.470 & 0.330 & 0.527 & 0.254 & 0.286 \\
			& RotNet-AT \cite{hendrycks2019selfsupervised} &  PGD & 0.333 & 0.501 & 0.507 & 0.361 & 0.551 & 0.314 & 0.315 \\
			& SOAP \cite{shi2021online} & & 0.504 & 0.608 & 0.398 & 0.606 & 0.425 & 0.522 & 0.533 \\
			& APAE \cite{goodge2020robustness} & & 0.104 & 0.155 & 0.240 & 0.202 & 0.229 & 0.191 & 0.278 \\
			& PrincipaLS (ours) & & \textbf{0.706} & \textbf{0.739} & \textbf{0.608} & \textbf{0.693} & \textbf{0.741} & \textbf{0.695} & \textbf{0.599} \\
			\noalign{\smallskip} \hline \noalign{\smallskip}
			& No Defense & Clean & 0.892 & 0.914 & 0.912 & 0.901 & 0.915 & 0.901 & 0.911 \\
			& No Defense & PGD & 0.088 & 0.223 & 0.152 & 0.177 & 0.177 & 0.262 & 0.438 \\
			\noalign{\smallskip} \cline{2-10} \noalign{\smallskip}
			& PGD-AT \cite{madry2018towards} & & 0.368 & 0.538 & 0.512 & 0.367 & 0.539 & 0.420 & 0.463 \\
			F-MNIST & FD \cite{Xie_2019_CVPR} & & 0.379 & 0.533 & 0.513 & 0.370 & 0.542 & 0.428 & 0.470 \\
			\cite{xiao2017fashion} & SAT \cite{xie2020smooth} & & 0.306 & 0.504 & 0.499 & 0.332 & 0.530 & 0.351 & 0.410 \\
			& RotNet-AT \cite{hendrycks2019selfsupervised} &  PGD & 0.375 & 0.542 & 0.509 & 0.365 & 0.524 & 0.396 & 0.429 \\
			& SOAP \cite{shi2021online} & & 0.475 & 0.509 & 0.313 & 0.477 & 0.386 & 0.548 & 0.521 \\
			& APAE \cite{goodge2020robustness} & & 0.174 & 0.366 & 0.300 & 0.246 & 0.398 & 0.310 & 0.409 \\
			& PrincipaLS (ours) & & \textbf{0.613} & \textbf{0.604} & \textbf{0.599} & \textbf{0.612} & \textbf{0.626} & \textbf{0.599} & \textbf{0.629} \\
			\noalign{\smallskip} \hline \noalign{\smallskip}
			& No Defense & Clean & 0.550 & 0.552 & 0.555 & 0.551 & 0.559 & 0.578 & 0.544 \\
			& No Defense & PGD & 0.034 & 0.073 & 0.051 & 0.037 & 0.027 & 0.087 & 0.141\\
			\noalign{\smallskip} \cline{2-10} \noalign{\smallskip}
			& PGD-AT \cite{madry2018towards} & & 0.145 & 0.177 & 0.195 & 0.146 & 0.182 & 0.157 & 0.167 \\
			CIFAR-10 & FD \cite{Xie_2019_CVPR} & & 0.147 & 0.180 & 0.206 & 0.150 & 0.187 & 0.152 & 0.170 \\
			\cite{krizhevsky2009learning} & SAT \cite{xie2020smooth} & & 0.141 & 0.170 & 0.186 & 0.141 & 0.181 & 0.107 & 0.160 \\
			& RotNet-AT \cite{hendrycks2019selfsupervised} &  PGD & 0.139 & 0.163 & 0.161 & 0.105 & 0.147 & 0.101 & 0.132 \\
			& SOAP \cite{shi2021online} & & 0.131 & 0.094 & 0.043 & 0.172 & 0.075 & 0.117 & 0.204 \\
			& APAE \cite{goodge2020robustness} & & 0.097 & 0.179 & 0.171 & 0.095 & 0.062 & 0.154 & 0.193 \\
			& PrincipaLS (ours) & & \textbf{0.246} & \textbf{0.247} & \textbf{0.252} & \textbf{0.244} & \textbf{0.242} & \textbf{0.245} & \textbf{0.248} \\	
			\noalign{\smallskip} \hline
		\end{tabular}
	\end{center}
\end{table*}

\subsection{Adversarial robustness} \label{sec:robustness}

\subsubsection{White-box robustness}
The robustness of one-class novelty detection against various white-box attacks is reported in Table~\ref{table:main_results_1}, where the vanilla AE is used. Without a defense, mAUROC scores drop significantly under all the white-box attacks, which shows the vulnerability of novelty detectors to the adversarial examples. PGD-AT improves adversarial robustness to a great extent. FD makes a slight improvement upon PGD-AT in most cases. SAT and Rot-AT seem not effective upon PGD-AT in the context of novelty detection. SOAP performs well in some cases but not uniformly. Compared to other methods, APAE generally shows less robustness. The proposed method, PrincipaLS, significantly increases mAUROC with PGD-AT, leading the other defenses by a decent margin. Moreover, PrincipaLS is consistently better across all the five white-box attacks, ranging from digital attacks to physically realizable attacks; on five datasets, ranging from toy datasets to realistic datasets, and from the image domain to video domain.

\noindent \textbf{PrincipaLS-knowledgeable attacks.}
As discussed above, in a white-box attack, attackers are aware of the presence of the defense mechanism, i.e., PrincipaLS (it is differentiable at inference time, see Sec.~\ref{sec:method}). However, they count on only the novelty detection objective (i.e., MSE loss, see Eq.~\eqref{mse_loss}) to generate adversarial examples. In this subsection, we follow the practice of the most recent adversarial defense studies such as \cite{shi2021online}, to thoroughly evaluate the proposed defense mechanism. More precisely, we try to find an adaptive attack \cite{papernot2017practical,tramer2020adaptive} by giving the full knowledge of the PrincipaLS defense mechanism to the attacker. We refer to this type of attack as \textit{PrincipaLS-knowledgeable attack}.

We construct two PrincipaLS-knowledgeable attacks, Knowledgeable A and Knowledgeable B, on top of the PGD attack. They jointly optimize Eq.~\eqref{mse_loss} and an auxiliary loss developed with the knowledge of PrincipaLS. Knowledgeable A attempts to minimize the $L_2$-norm between the latent space before and after the PrincipaLS transformation. The intuition is to void PrincipaLS such that the input and the output latent space of PrincipaLS become closer. In other words, Knowledgeable A replaces Eq.~\eqref{mse_loss} with the following equation:
\begin{equation}
\label{adaptive_1}
\mathcal{L} = y \parallel \hat{\mathbf{X}}^t - \mathbf{X}^t \parallel_2 - \lambda_A \parallel \mathbf{Z}^t_{PrincipaLS} - \mathbf{Z}_{adv}^t \parallel_2, 
\end{equation}
where, $\lambda_A$ is a trade-off parameter. Knowledgeable B attempts to maximize the $L_2$-norm between the latent space of the current adversarial example $\mathbf{X}^t$ and its clean counterpart $\mathbf{X}^0$ after the PrincipaLS transformation. The intuition is to keep the adversarial latent space away from the clean one. In other words, Knowledgeable B replaces Eq.~\eqref{mse_loss} with the following equation:
\begin{equation}
\label{adaptive_2}
\mathcal{L} = y \parallel \hat{\mathbf{X}}^t - \mathbf{X}^t \parallel_2 + \lambda_B \parallel \mathbf{Z}^t_{PrincipaLS} - \mathbf{Z}^0_{PrincipaLS} \parallel_2, 
\end{equation}
where, $\lambda_B$ is a trade-off parameter. When $\lambda_A=0$ or $\lambda_B=0$, the PrincipaLS-knowledgeable attacks reduce to the conventional white-box attacks.

\begin{figure}[!t]
	\centering
	\includegraphics[width=0.48\textwidth]{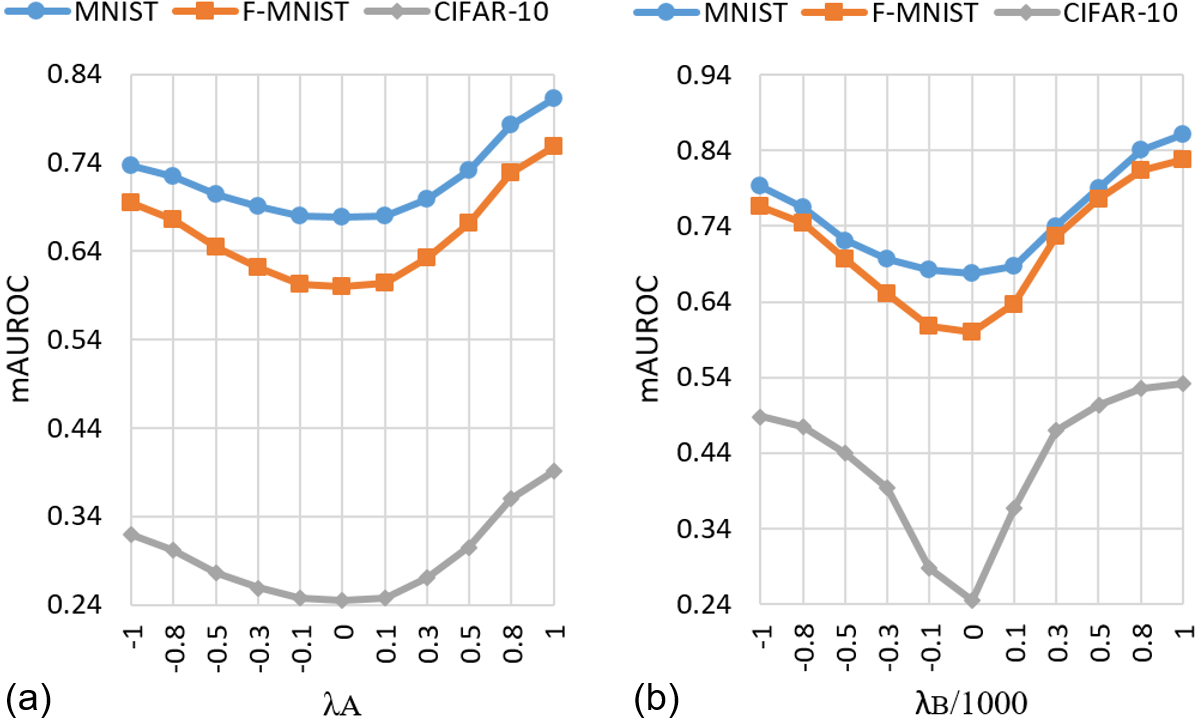}
	\caption{The mAUROC of PrincipaLS under PrincipaLS-knowledgeable attacks with varied trade-off parameters. (a) Knowledgeable A. (b) Knowledgeable B.}
	\label{fig:adaptive}
\end{figure}

In Fig.~\ref{fig:adaptive}, we can observe that mAUROC monotonously increases as $|\lambda_A|$ or $|\lambda_B|$ increases. That is, these PrincipaLS-knowledgeable attacks cannot further reduce PrincipaLS's mAUROC, and the additional auxiliary loss terms would attenuate the MSE loss gradients. This indicates that attackers cannot straightforwardly benefit from the knowledge of PrincipaLS. Hence, the conventional white-box attack still has the greatest attacking strength. This result shows that it is not easy to find a stronger attack to break PrincipaLS, even with the full knowledge of the PrincipaLS mechanism.

% Adaptive attacks

\subsubsection{Black-box robustness}
The robustness against black-box attacks \cite{papernot2017practical} is shown in the second last column of Table~\ref{table:main_results_1}. Here, we consider a naturally trained (i.e., train with only clean data) GPND as a substitute model and apply MI-FGSM, which has better transferability, to generate black-box adversarial examples for target models. As we can see, the defenses with PGD-AT degrade black-box robustness, which is identical to the observation in classification tasks \cite{tramer2018ensemble}. SOAP, which is without using AT, shows better black-box robustness. PrincipaLS greatly improves the black-box robustness on most datasets even with PGD-AT. The naturally trained PrincipaLS model achieves 0.907, 0.742 and 0.332 mAUROC on MNIST, F-MNIST and CIFAR-10, respectively, under the black-box attack.

\subsubsection{Generalizability}
Table~\ref{table:main_results_2} shows the adversarial robustness of various state-of-the-art novelty detection models. All of them are susceptible to adversarial attacks. We attach the PrincipaLS module to these models to protect them. We can see that PrincipaLS uniformly robustifies all of these novelty detectors and significantly outperforms the other defense approaches. This confirms that PrincipaLS can be applied to a wide variety of the present novelty detection methods, demonstrating its excellent generalizability. Detailed class-wise AUROC scores can be found in Sec.~\ref{sec:auroc} in Supplementary.

% Clean performance
\setlength{\tabcolsep}{7pt}
\begin{table}
	\begin{center}
		\caption{The mAUROC of models under clean data.}
		\label{table:clean_performance}
		\begin{tabular}{r | ccc}
			\hline \noalign{\smallskip} \noalign{\smallskip}
			Defense & MNIST & F-MNIST & CIFAR-10 \\
			\hline \noalign{\smallskip} \noalign{\smallskip}
			No Defense & 0.964 & 0.892 & 0.550 \\
			FD \cite{Xie_2019_CVPR} & 0.965 & 0.892 & 0.551 \\
			SAT \cite{xie2020smooth} & 0.949 & 0.883 & 0.543 \\
			RotNet-AT \cite{hendrycks2019selfsupervised} &  0.963 & 0.897 & 0.554 \\
			SOAP \cite{shi2021online} & 0.940 & 0.876 & 0.546 \\
			APAE \cite{goodge2020robustness} & 0.925 & 0.861 & 0.552 \\
			PrincipaLS (ours) & \textbf{0.973} & \textbf{0.922} & \textbf{0.578} \\
			\noalign{\smallskip} \hline
		\end{tabular}
	\end{center}
\end{table}

% Inference speed
\setlength{\tabcolsep}{7pt}
\begin{table}
	\begin{center}
		\caption{The inference speed of each defense. The test images are from CIFAR-10 with an input size of 32 $\times$ 32. The experiment is performed on a single NVIDIA RTX 2080 Ti GPU.}
		\label{table:inference_speed}
		\begin{tabular}{r | rr}
			\hline \noalign{\smallskip} \noalign{\smallskip}
			Defense & Speed (FPS) & Difference \\
			\hline \noalign{\smallskip} \noalign{\smallskip}
			No Defense & 18.0 $\times 10^3$ & \\
			FD \cite{Xie_2019_CVPR} & 6.8 $\times 10^3$ & -62.2\%\\
			SAT \cite{xie2020smooth} & 18.2 $\times 10^3$ & +1.1\% \\
			RotNet-AT \cite{hendrycks2019selfsupervised} & 18.0 $\times 10^3$ & -0.0\% \\
			SOAP \cite{shi2021online} & 3.1 $\times 10^3$ & -82.2\% \\
			APAE \cite{goodge2020robustness} & 4.0 $\times 10^3$ & -77.8\% \\
			PrincipaLS (ours) & 15.6 $\times 10^3$ & -13.3\% \\
			\noalign{\smallskip} \hline
		\end{tabular}
	\end{center}
\end{table}

\subsection{Performance on clean data}  \label{sec:clean_performance}
We also evaluate the performance of PrincipaLS on clean data. In this experiment, all the models are naturally trained. As shown in Table~\ref{table:clean_performance}, PrincipaLS improves the performance upon the original network architecture (No Defense), while, the other defenses do not make obvious improvements. This shows that PrincipaLS generalizes better for both clean data and adversarial examples. PrincipaLS enjoys this benefit because the principal latent components are learned from only the latent space of the known class. Due to this, when transforming the latent space of any novel class image, PrincipaLS projects it into the known class space defined by the principal latent component. This brings the transformed latent space closer to the latent space of the known class, resulting in the decoder trying to reconstruct it into a known class image. Subsequently, this produces high reconstruction error for the novel class images while barely affecting the reconstruction of the known class images.
%Hence, the model performance with PRL is better for clean images compared to original original network architecture (No Defense).

\subsection{Inference speed}
The PrincipaLS module is light-weight and computationally efficient. We test the inference speed of each defense via images are from CIFAR-10 with an input size of 32 $\times$ 32. The experiment is performed on a single NVIDIA RTX 2080 Ti GPU. As can be seen in Table~\ref{table:inference_speed}, when the PrincipaLS module is attached to an AE, the inference speed only decreases 13.3\%. This cost turns to significant improvements in robustness. We use a compact AE architecture as described in Sec.~\ref{sec51}. If a deeper AE architecture is considered, PrincipaLS's relative computational overhead will be even lower. In contrast, FD contains a heavy feature denoising block which decreases inference speed by 62.2\%. SOAP and APAE rely on adversarial purification processes at inference time, greatly increasing computational cost.

\subsection{Analysis} \label{sec_ablation}

\subsubsection{Ablation study}
Table~\ref{table:ablation} reports the results of different PrincipaLS variants. First, Vector-PCA alone significantly improves the robustness upon PGD-AT. This shows that the mechanism of replacing perturbed latent vectors by the incrementally-trained principal latent vector is effective. As discussed earlier, in PrincipaLS the adversaries can stay only on the scaling factors of the principal latent vector. Next, we further remove the adversaries with the help of denoising operation on the spatial dimension. We try to deploy a feature denoising block \cite{Xie_2019_CVPR} after the forward Vector-PCA. This baseline is denoted as \textit{Vector-PCA+FD}. This makes a slight improvement over Vector-PCA baseline. Finally, the complete PrincipaLS uses Spatial-PCA for this purpose instead, achieveing great mAUROC increase. This shows Spatial-PCA's advantage over FD in our case.

% Ablation study
\setlength{\tabcolsep}{7pt}
\begin{table}
	\begin{center}
		\caption{The mAUROC of different PrincipaLS variants under PGD attack.}
		\label{table:ablation}
		\begin{tabular}{r | ccc}
			\hline \noalign{\smallskip} \noalign{\smallskip}
			Defense & MNIST & F-MNIST & CIFAR-10 \\
			\noalign{\smallskip} \hline \noalign{\smallskip}
			PGD-AT \cite{madry2018towards} & 0.357 & 0.368 & 0.145 \\
			Vector-PCA & 0.566 & 0.499 & 0.215 \\
			Vector-PCA+FD & 0.582 & 0.505 & 0.215 \\
			PrincipaLS (ours) & \textbf{0.706} & \textbf{0.613} & \textbf{0.246} \\
			\noalign{\smallskip} \hline
		\end{tabular}
	\end{center}
\end{table}

% Trade-off
\setlength{\tabcolsep}{5.5pt}
\begin{table}[!t]
	\begin{center}
		\caption{The trade-off analysis of PrincipaLS's $k_V$ and $k_S$ values on MNIST dataset.}
		\label{table:tradeoff}
		\begin{tabular}{c | c | cccc}
			\hline \noalign{\smallskip} \noalign{\smallskip}
			Input & Original AE & $k_V=1$ & $k_V=4$ & $k_V=16$ & $k_V=64$ \\
			\hline \noalign{\smallskip} \noalign{\smallskip}
			Clean & 0.964 & 0.973 & 0.975 & 0.971 & 0.971 \\
			PGD & 0.357 & 0.706 & 0.621 & 0.581 & 0.557 \\
			\hline \noalign{\smallskip} \noalign{\smallskip}
			Input & Vec-PCA only & $k_S=1$ & $k_S=4$ & $k_S=8$ & $k_S=12$ \\
			\hline \noalign{\smallskip} \noalign{\smallskip}
			Clean & 0.968 & 0.937 & 0.951 & 0.973 & 0.973 \\
			PGD & 0.566 & 0.549 & 0.681 & 0.706 & 0.667 \\
			\noalign{\smallskip} \hline
		\end{tabular}
	\end{center}
\end{table}

\subsubsection{Trade-off of $k_V$ and $k_S$ values}
We look into the trade-off of PrincipaLS's $k_V$ and $k_S$ values. Table~\ref{table:tradeoff} reports the results on the MNIST dataset. For both varying $k_V$ (fix $k_S$=8) and $k_S$ (fix $k_V$=1), we observe that larger $k$ leads to lower PGD accuracy but higher clean accuracy in general. The reason is that using larger $k$ retains more semantic information of feature maps while keeping more adversaries simultaneously. $k_S$=1 is an exception. It has lower PGD accuracy because it loses too much information. According to this trade-off analysis, we set $k_V$=1 and $k_S$=8 for PrincipaLS as discussed in Sec.~\ref{sec51}.

\subsubsection{Attack budgets}
\begin{figure}[!t]
	\centering
	\includegraphics[width=0.48\textwidth]{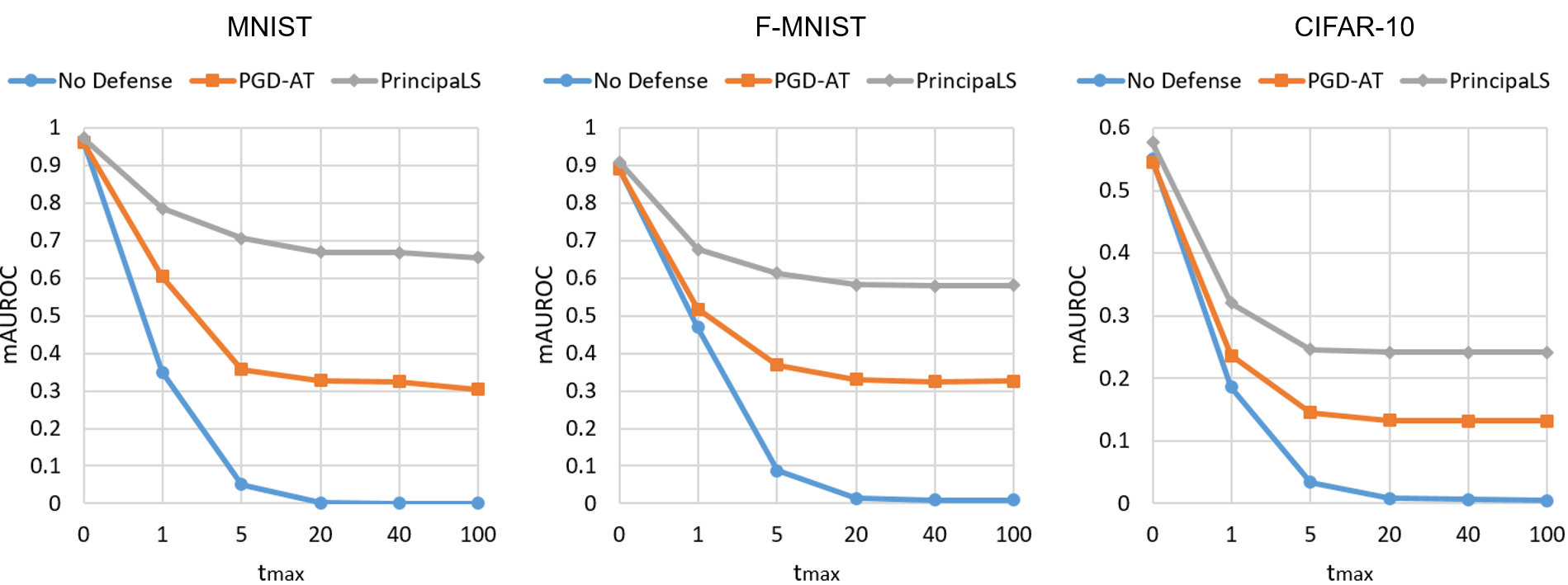}
	%\captionsetup{justification=centering}
	\caption{The mAUROC of models under PGD attack with varied numbers of attack iterations $t_{max}$.}
	\label{fig:t_max}
\end{figure}

\begin{figure}[!t]
	\centering
	\includegraphics[width=0.48\textwidth]{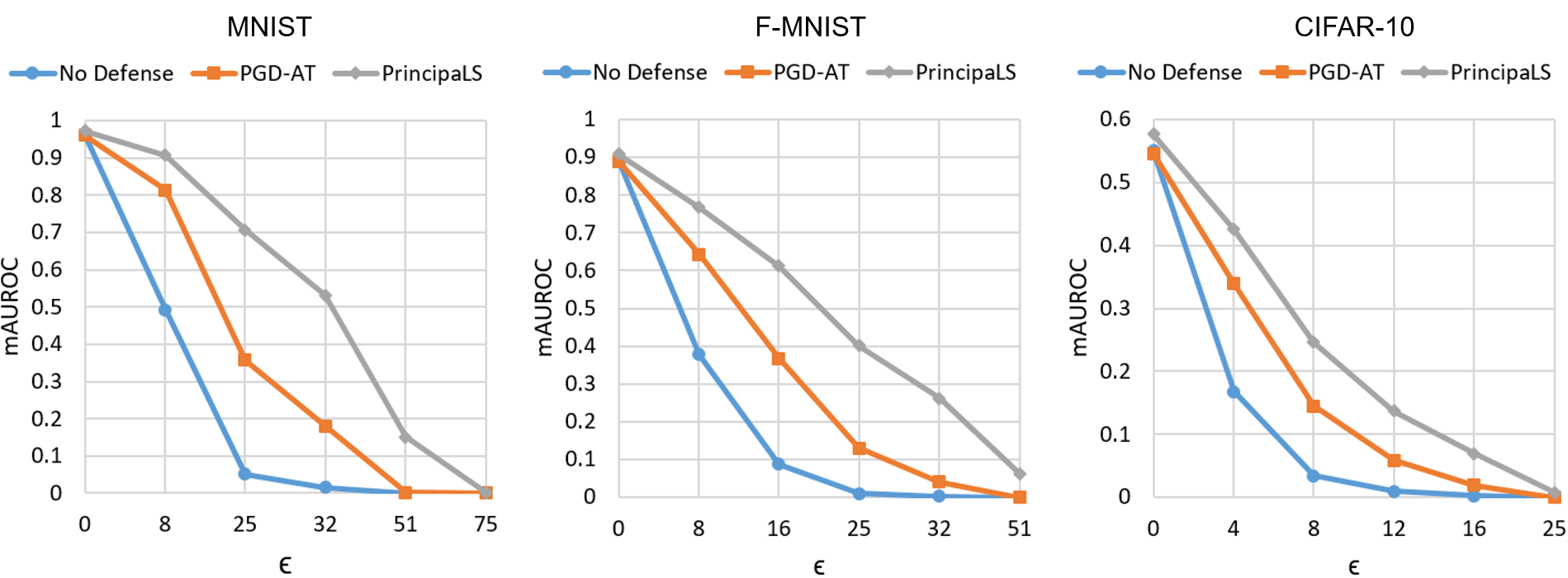}
	%\captionsetup{justification=centering}
	\caption{The mAUROC of models under PGD attack with varied perturbation sizes $\epsilon$.}
	\label{fig:epsilon}
\end{figure}

To fully evaluate the effectiveness of the proposed PrincipaLS, we test its scalability to different attack budgets. We vary the attack budgets by two aspects: The number of attack iterations $t_{max}$ and perturbation sizes $\epsilon$. The results are presented in Fig.~\ref{fig:t_max} and Fig.~\ref{fig:epsilon}, respectively.

First, we can see that the attack strength does not increase obviously along with the increase of $t_{max}$. This observation is consistent with that of Madry et al. \cite{madry2018towards} and Xie et al. \cite{Xie_2019_CVPR}. The proposed PrincipaLS shows constant adversarial robustness and consistently performs better than No Defense and PGD-AT under different $t_{max}$. On the other hand, the attack strength significantly increases along with the increase of $\epsilon$. It can be observed that PrincipaLS consistently demonstrates better robustness under different $\epsilon$. Apparently, PrincipaLS is scalable to different attack budgets.

\subsubsection{Stability of latent space}
\begin{figure}[!t]
	\centering
	\includegraphics[width=0.48\textwidth]{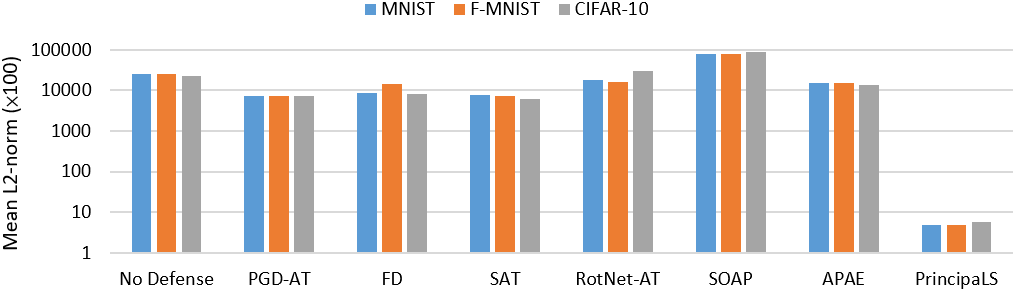}
	\caption{Mean $L_2$-norm between the latent space of PGD adversarial examples and that of their clean counterpart on different defenses. The values are the mean over an entire dataset.}
	\label{fig:feature_diff}
\end{figure}

%As discussed in Sec. \ref{sec:robustness}, We analyze the stability of different defense methods' latent space to show why PrincipaLS works.
We compute the mean $L_2$-norm between the latent space of adversarial examples and that of their clean counterpart: $\parallel \mathbf{Z}_{adv} - \mathbf{Z} \parallel_2$. As can be seen in Fig.~\ref{fig:feature_diff}, PrincipaLS's mean $L_2$-norm is three orders of magnitude smaller than the other defenses. This indicates that PrincipaLS's latent space are barely affected by adversaries, showing PrincipaLS's effectiveness in adversary removal.
%Such excellent stability makes PrincipaLS robust.

% t_max

% epsilon

% Feature differences

% Histogram
\begin{figure}[!t]
	\centering
	\includegraphics[width=0.48\textwidth]{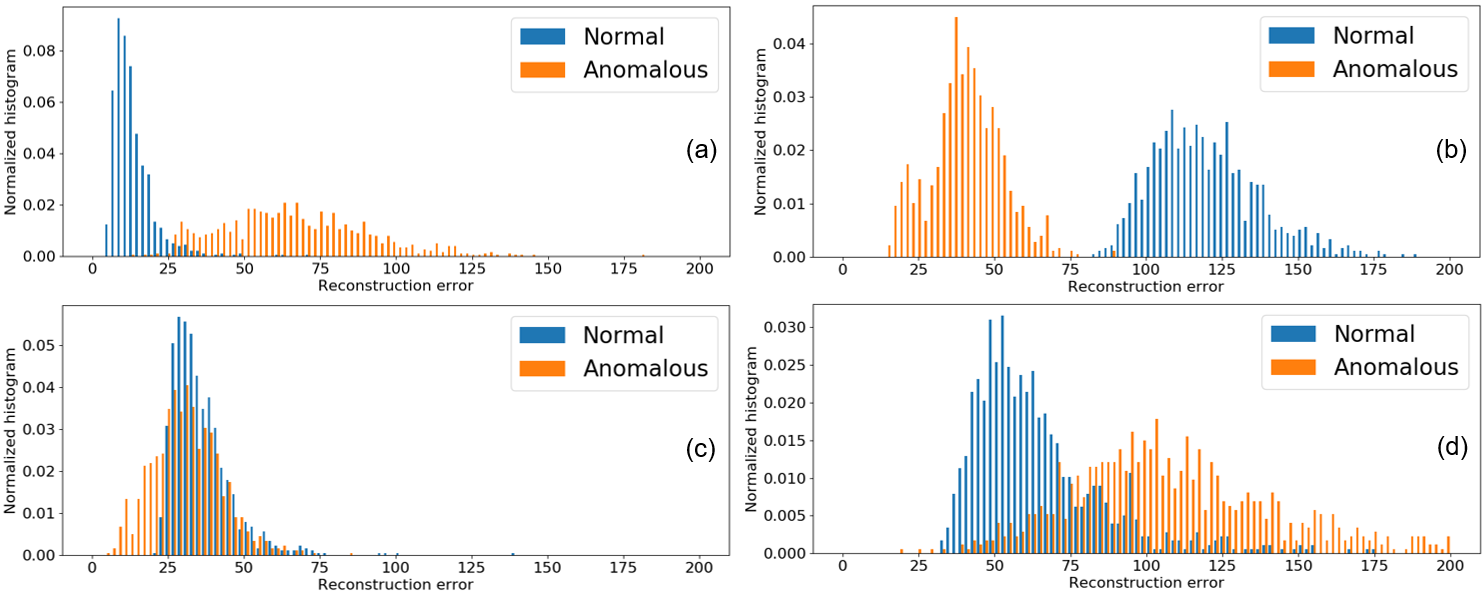}
	\caption{Histograms of reconstruction errors. (a) No Defense under clean data. (b) No Defense under PGD attack. (c) PGD-AT under PGD attack. (d) PrincipaLS under PGD attack. Digit 0 of MNIST is set to normal data, and the other digits are anomalous.}
	\label{fig:histogram}
\end{figure}

\subsubsection{Reconstruction errors}
For an AE-style novelty detection model, normal data and anomalous data are expected to get low and high reconstruction errors, respectively. The model follows this behavior given clean data, as shown in Fig.~\ref{fig:histogram}(a). When an attacker attempts to maximize the reconstruction errors of normal data and minimize that of anomalous data, the model would make wrong predictions, shown in Fig.~\ref{fig:histogram}(b). Fig.~\ref{fig:histogram}(c) shows that PGD-AT pulls back the enlarged reconstruction errors of normal data, but they still overlap for the anomalous data. In Fig.~\ref{fig:histogram}, it can be observed that PrincipaLS pushes the reconstruction errors of anomalous data with better margin. Although the reconstruction errors of normal data also increases, the gap between normal and anomalous data is retained resulting in PrincipaLS performing better under attacks.

\subsubsection{Reconstructed images}
\begin{figure}[!t]
	\centering
	\includegraphics[width=0.48\textwidth]{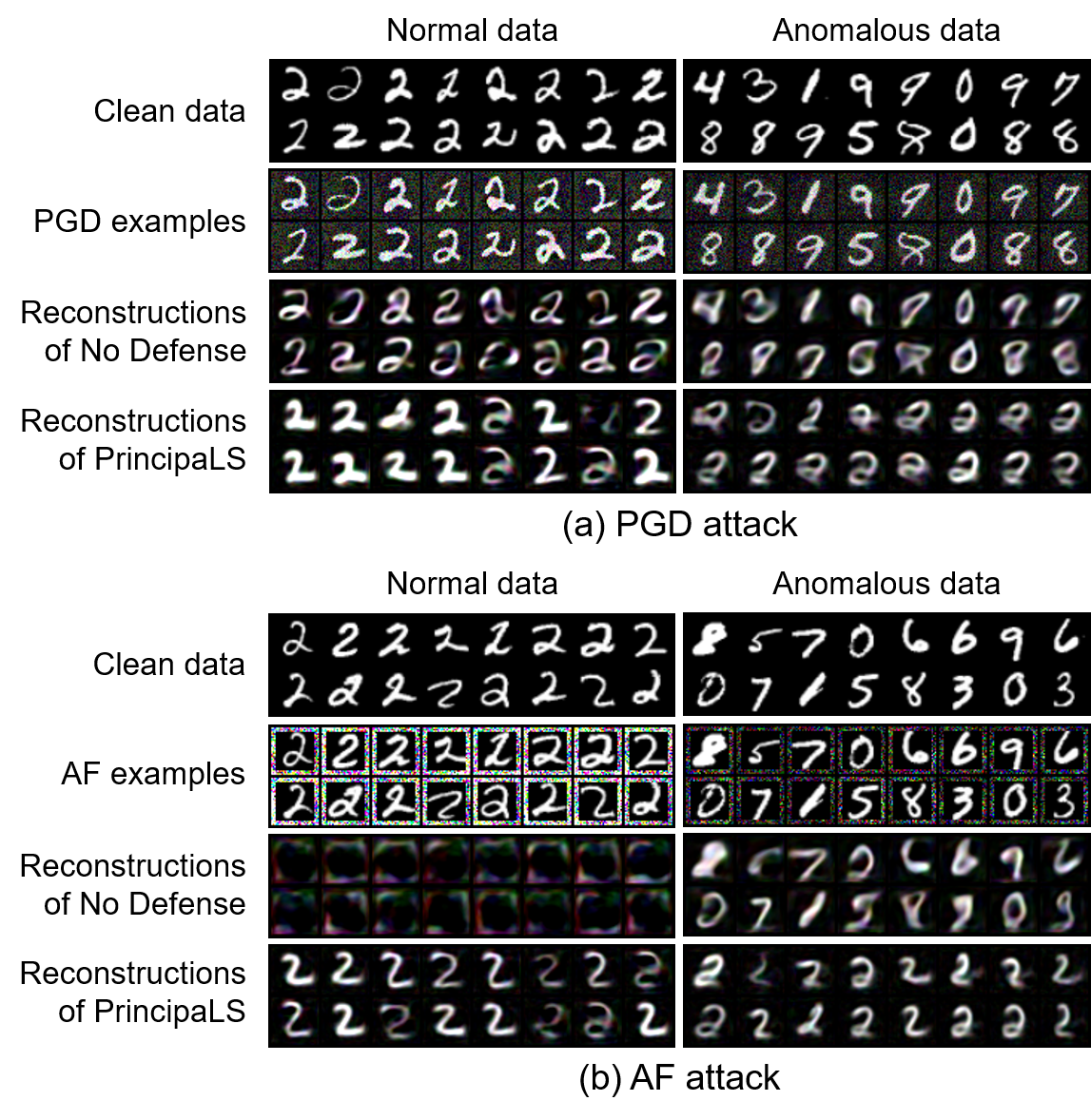}
	\caption{Reconstructions under (a) PGD attack with $\epsilon = 76/255$ and (b) AF attack with framing with $= 1$, $\epsilon = 255/255$. Digit 2 is set to normal data, and the other digits are anomalous.}
	\label{fig:visual}
\end{figure}

Fig.~\ref{fig:visual} compares the reconstructed images of No Defense model and PrincipaLS under PGD and AF attacks. Digit 2 of MNIST is used as the known class. In the PGD case, No Defense model produces decent reconstructions for both adversarial normal and anomalous data. Hence, the reconstruction error gap between normal data and anomalous data is insufficiently large. In the AF case, No Defense model still captures the shape of the adversarial anomalous data and thus produces fair reconstructions, but it fails to reconstruct recognizable patterns for adversarial normal data. Therefore, the resulting reconstruction errors would cause wrong predictions. Such observations are consistent with the quantitative results that it is not adversarially robust. In contrast, PrincipaLS reconstructs every data into the known class of digit 2. Hence, even under attacks, PrincipaLS can obtain very high reconstruction errors from anomalous data and low errors from normal data.

% Visual results

% Attack
\setlength{\tabcolsep}{4.5pt}
\begin{table}
	\begin{center}
		\caption{The mAUROC of models under PGD, PGD-normal, PGD-latent, PGD-clean and PGD-anomalous attacks. Underlines denote the lowest mAUROC, which indicates the strongest attack.}
		\label{table:attack}
		\begin{tabular}{r | c | ccc}
			\hline \noalign{\smallskip} \noalign{\smallskip}
			Defense & Attack method & MNIST & F-MNIST & CIFAR-10 \\
			\noalign{\smallskip} \hline \noalign{\smallskip}
			& Clean & 0.964 & 0.892 & 0.550 \\
			\noalign{\smallskip} \cline{2-5} \noalign{\smallskip}
			& PGD & \underline{0.051} & \underline{0.088} & \underline{0.034} \\
			No Defense & PGD-normal & 0.167 & 0.284 & 0.111 \\
			& PGD-latent & 0.773 & 0.715 & 0.433 \\
			& PGD-clean & 0.106 & 0.180 & 0.070 \\
			& PGD-anomalous & 0.939 & 0.788 & 0.332 \\
			\noalign{\smallskip} \hline \noalign{\smallskip}
			& PGD & \underline{0.357} & \underline{0.368} & \underline{0.145} \\
			& PGD-normal & 0.745 & 0.656 & 0.309 \\
			PGD-AT \cite{madry2018towards} & PGD-latent & 0.914 & 0.784 & 0.448 \\
			& PGD-clean & 0.863 & 0.802 & 0.403 \\
			& PGD-anomalous & 0.753 & 0.677 & 0.328 \\
			\noalign{\smallskip} \hline \noalign{\smallskip}
			& PGD & \underline{0.366} & \underline{0.379} & \underline{0.147} \\
			& PGD-normal & 0.750 & 0.654 & 0.309 \\
			FD \cite{Xie_2019_CVPR} & PGD-latent & 0.906 & 0.762 & 0.447 \\
			& PGD-clean & 0.871 & 0.794 & 0.401 \\
			& PGD-anomalous & 0.761 & 0.673 & 0.331 \\
			\noalign{\smallskip} \hline \noalign{\smallskip}
			& PGD & \underline{0.706} & \underline{0.613} & \underline{0.246} \\
			& PGD-normal & 0.905 & 0.786 & 0.399 \\
			PrincipaLS (ours) & PGD-latent & 0.962 & 0.882 & 0.547 \\
			& PGD-clean & 0.936 & 0.867 & 0.520 \\
			& PGD-anomalous & 0.881 & 0.781 & 0.407 \\
			\noalign{\smallskip} \hline
		\end{tabular}
	\end{center}
\end{table}

\section{Discussion}

\subsection{Further comparison with ARAE} \label{sec:arae}
ARAE \cite{salehi2021arae} somewhat refers to the adversarial robustness of novelty detection though its main purpose is improving performance. As mentioned in Sec.~\ref{sec3}, ARAE's adversarial robustness is not thoroughly evaluated. In this section, we make a comprehensive comparison with ARAE.

First, ARAE evaluates adversarial robustness by crafting adversarial examples from only the normal test data (the known class). We refer this attack as \textit{PGD-normal}. Instead, our attack method crafts adversarial examples from every test data regardless of their class (see Sec.~\ref{sec3}). We reproduce PGD-normal with the same setting as in Sec.~\ref{sec51}. As shown in Table~\ref{table:attack}, the proposed attack (denoted as PGD) is stronger than PGD-normal, in which PGD obtains lower mAUROC across all the considered defense methods and datasets. It is intuitive that perturbing every input data poses a stronger attack.

Second, ARAE performes AT on the latent space-based adversarial examples. We name this attack as \textit{PGD-latent}. Instead, in this paper, we perform AT on the output space-based adversarial examples (see Sec.~\ref{sec3}). We reproduce PGD-latent with the same setting as in Sec.~\ref{sec51}. Specifically, PGD-latent replaces the loss objective Eq.~\eqref{mse_loss} with follows:
% PGD-latent
\begin{equation}  \label{pgd_latent}
\mathcal{L}(\mathbf{X}^t, \mathbf{X}, y) = y \parallel Enc(\mathbf{X}^t) - Enc(\mathbf{X}) \parallel_2,
\end{equation}
where $Enc$ denotes the encoder in an AE. As can be seen in Table~\ref{table:attack}, PGD is much stronger than PGD-latent, in which PGD obtains lower mAUROC across all the considered defense methods and datasets. Therefore, we perform AT by minimizing Eq.~\eqref{mse_loss} to make a stronger defense.

Third, a novelty detector would not know whether an input image is adversarial or not during inference. In other words, if the given input is an adversarial image, the clean counterpart is not available at test time. Hence, a novelty detector should compute the novelty score by the reconstruction error between the reconstructed image and the ``input image" (regardless it is clean or adversarial) instead of that between the reconstructed image and the ``clean image". For instance, if a given test image is an adversarial example $\mathbf{X}_{adv}$, a novelty detector should compute $\parallel \hat{\mathbf{X}}_{adv} - \mathbf{X}_{adv} \parallel_2$ instead of $\parallel \hat{\mathbf{X}}_{adv} - \mathbf{X} \parallel_2$ as the novelty score, where $\mathbf{X}$ is the clean image. Therefore, to craft a strong adversarial example, one should maximize the reconstruction error between the reconstructed image and the ``input image" (regardless it is clean or adversarial). The proposed attack is based on this nature; that is, at each attack iteration, we maximize the reconstruction error between the current adversarial example and the reconstruction of that current adversarial example (see Eq.~\eqref{mse_loss}). We make an attack variant, \textit{PGD-clean}, which maximizes the reconstruction error between the clean image and the reconstruction of the current adversarial example. Specifically, PGD-clean replaces the loss objective Eq.~\eqref{mse_loss} with follows:
% PGD-clean
\begin{equation}  \label{pgd_clean}
\mathcal{L}(\hat{\mathbf{X}}^t, \mathbf{X}, y) = y \parallel \hat{\mathbf{X}}^t - \mathbf{X} \parallel_2.
\end{equation}
ARAE uses this form. As shown in Table~\ref{table:attack}, PGD is much stronger than PGD-clean, in which PGD obtains lower mAUROC across all the considered defense methods and datasets. Therefore, we perform AT by minimizing Eq.~\eqref{mse_loss} to make a stronger defense.

In summary, the proposed attack is much stronger than PGD-normal, PGD-latent and PGD-clean. Hence, we can carefully and strictly evaluate the adversarial robustness of novelty detectors. Moreover, conducting AT on a stronger attack can enhance robustness to a greater extent, so using the proposed attack for doing AT can make novelty detectors much more robust. We hope to provide researchers with a solid benchmark for future work on the adversarial robustness of novelty detection.

\subsection{Further comparison with APAE} \label{sec:apae}
To the best of our knowledge, APAE \cite{goodge2020robustness} might be the only present defense designed for anomaly detection. However, as mentioned in Sec.~\ref{sec3}, APAE's adversarial robustness is not thoroughly evaluated. In this section, we make more comparisons with APAE.

First, APAE evaluates adversarial robustness by crafting adversarial examples from only the anomalous test data (the unknown classes), which is contrary to ARAE's PGD-normal (Sec.~\ref{sec:arae}). We name this attack \textit{PGD-anomalous}. Instead, our attack method crafts adversarial examples from every test data regardless of their class (see Sec.~\ref{sec3}). We reproduce PGD-anomalous with the same setting as in Sec.~\ref{sec51}. As shown in Table~\ref{table:attack}, the proposed attack (denoted as PGD) is stronger than PGD-anomalous, in which PGD obtains lower mAUROC across all the considered defense methods and datasets. It is intuitive that perturbing every input data poses a stronger attack. On the other hand, No Defense attains the best mAUROC compared with the other defenses. The reason is that these defenses use only normal data to do AT, so they overfit the adversarial normal data and show less robustness against PGD-anomalous.

Second, APAE claims that AT does not apply to the novelty detection problem. In contrast, in this paper, we demonstrate that AT actually does apply to novelty detection, in which we can craft adversarial examples for the normal data and use them to train the target model. Indeed, using AT is less robust to PGD-anomalous as shown in Table~\ref{table:attack}. However, for the stronger attacks (i.e., the proposed attack) that contain adversarial normal data, AT can significantly improve the robustness.

Apparently, we construct a more appropriate evaluation protocol to fully test the adversarial robustness of novelty detectors. With a proper evaluation protocol, we are able to design a much better defense method accordingly.

\subsection{Does PrincipaLS work on image classification?}
The proposed PrincipaLS method is specifically designed for the novelty detection task. As discussed in Sec.~\ref{sec:introduction} and Sec.~\ref{sec:method}, it leverages the task-specific knowledge that novelty detectors are only required to retain information about normal data, thereby resulting in the preferred high reconstruction errors for anomalous data. This property allows PrincipaLS to largely manipulate the latent space of novelty detectors to remove adversaries, while maintaining the performance on clean data. It can be noted that such property is unique to novelty detection, as most visual recognition problems (e.g., image classification) require a model containing high-level semantic information. Hence, a large manipulation on the latent space would limit the model capacity and thus degrade accuracy.

% Image classification
\setlength{\tabcolsep}{10pt}
\begin{table}[t!]
	\begin{center}
		\caption{Image classification accuracy (\%) on CIFAR-10.}
		\label{table:classifier}
		\begin{tabular}{r | c | c}
			\hline \noalign{\smallskip} \noalign{\smallskip}
			Defense & Clean & PGD \\
			\noalign{\smallskip} \hline \noalign{\smallskip}
			No Defense & 94.0 & 0.0 \\
			PGD-AT \cite{madry2018towards} & 83.3 & 51.2 \\
			FD \cite{Xie_2019_CVPR} & 83.3 & 51.5 \\
			\noalign{\smallskip} \hline \noalign{\smallskip}
			PrincipaLS $k_V=1$ & 36.0 & 30.6 \\
			PrincipaLS $k_V=8$ & 71.1 & 47.0 \\
			PrincipaLS $k_V=64$ & 72.7 & 48.3 \\
			\noalign{\smallskip} \hline
		\end{tabular}
	\end{center}
\end{table}

% Multiple class novelty detection
\setlength{\tabcolsep}{10pt}
\begin{table}[t!]
	\begin{center}
		\caption{The AUROC of multiple class novelty detection on MNIST. Digit 0 and digit 2 are set to the known classes.}
		\label{table:mcnd}
		\begin{tabular}{r | c | c}
			\hline \noalign{\smallskip} \noalign{\smallskip}
			Defense & Clean & PGD \\
			\noalign{\smallskip} \hline \noalign{\smallskip}
			No Defense & 0.926 & 0.051 \\
			PGD-AT \cite{madry2018towards} & 0.926 & 0.052 \\
			FD \cite{Xie_2019_CVPR} & 0.926 & 0.066 \\
			PrincipaLS (ours) & \textbf{0.954} & \textbf{0.412} \\
			\noalign{\smallskip} \hline
		\end{tabular}
	\end{center}
\end{table}

% Visual results mult
\begin{figure}[!t]
	\centering
	\includegraphics[width=0.48\textwidth]{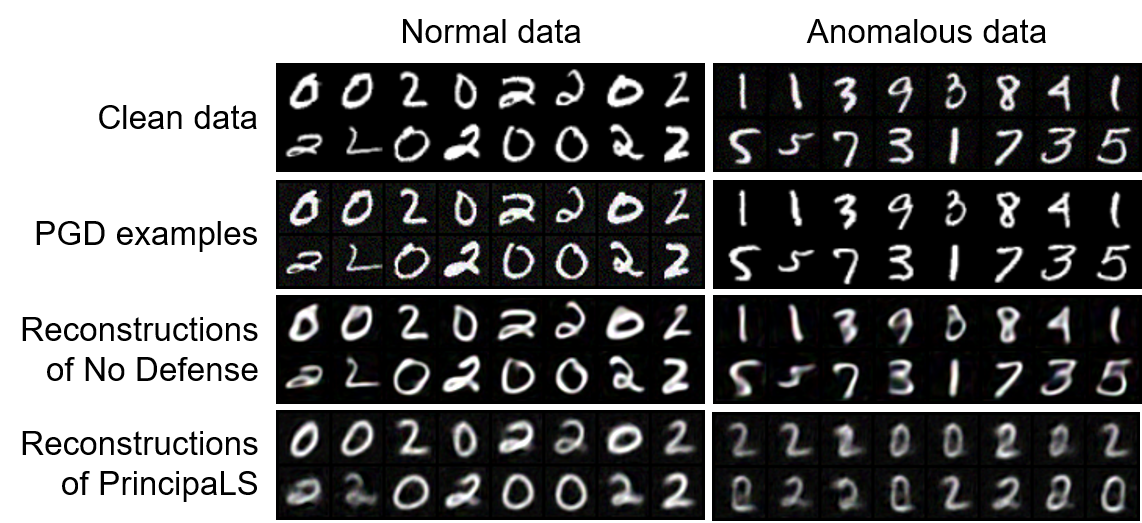}
	\caption{Reconstructions under PGD attack with $\epsilon = 25/255$. Digit 0 and digit 2 are set to normal data, and the other digits are anomalous.}
	\label{fig:visual_mult}
\end{figure}

To demonstrate this, we apply PrincipaLS to the image classification task on CIFAR-10 \cite{krizhevsky2009learning}. We attach the PrincipaLS module to ResNet-18 \cite{he2016deep} between the last convolutional layer and the fully-connected layer. We use PGD attack with $\epsilon = 8/255$ for testing and AT. Table~\ref{table:classifier} shows that PGD-AT and FD, which are originally designed for image classification, effectively improve adversarial robustness. As expected, PrincipaLS obtains both lower clean and PGD accuracies. The reason is that the PrincipaLS operation reduces the model capacity of learning high-level semantic representations, making the latent space insufficiently discriminative for classification. We can see that larger $k_V$ achieves higher PGD accuracy, which is an opposite trend to that in novelty detection (see Table~\ref{table:tradeoff}). In other words, different from novelty detectors, image classifiers cannot enjoy the principal latent space since it loses too much semantic information.

PrincipaLS, designed for novelty detection, does not work on image classification; conversely, the defenses designed for image classification are not that effective on novelty detection as shown in Table~\ref{table:main_results_1} and Table~\ref{table:main_results_2}. Apparently, these two vision tasks have different characteristics and thus need different adversarial defenses. This demonstrates the need for a defense method specifically designed for novelty detection and thus highlights the contribution of this paper.

\subsection{Applying to multiple class novelty detection}
Multiple class novelty detection \cite{oza2020utilizing,perera2019deep} has the same problem setting as one-class novelty detection except that it considers multiple known classes. It is more challenging, as it needs to characterize the underlying distributions of multiple known classes and identify novel classes given such knowledge of multiple known classes. We explore applying the proposed PrincipaLS to multiple class novelty detection.

In this experiment, we define digit 0 and digit 2 of MNIST dataset as the known classes (normal data), and the rest of the digits are novel classes (anomalous data). In Table~\ref{table:mcnd}, we can find that PGD-AT and FD do not improve adversarial robustness, which demonstrates that image classification-based defense approaches cannot protect multiple class novelty detectors. In contrast, PrincipaLS significantly increases PGD accuracy, showing its potential for applying to multiple class novelty detection. Fig.~\ref{fig:visual_mult} shows that PrincipaLS reconstructs every normal and anomalous data into the known classes of digit 0 or digit 2. Therefore, even under adversarial attacks, PrincipaLS can obtain very high reconstruction errors from anomalous data and low errors from normal data. Furthermore, PrincipaLS achieves higher clean accuracy, as the principal latent components can exclusively characterize the latent space of the known classes (see Sec.~\ref{sec:clean_performance}).
%Note that this is just a simple experiment to see the possibility of applying PrincipaLS to multiple class novelty detection. We will leave a more complete evaluation with future work.

\section{Conclusion}
In this paper, we study the adversarial robustness in the context of one-class novelty detection problem. We show that existing novelty detection models are vulnerable to adversarial perturbations and then propose a defense method referred to as Principal Latent Space (PrincipaLS). Specifically, PrincipaLS purifies the latent space by the incrementally-trained cascade PCA process. Moreover, we construct a generic evaluation framework to fully test the effectiveness of the proposed PrincipaLS. We perform extensive experiments on multiple datasets with multiple existing novelty detection models and consider various attacks to show that PrincipaLS consistently improves adversarial robustness.

% or

\ifCLASSOPTIONcompsoc
  \section*{Acknowledgments}
\else
  \section*{Acknowledgment}
\fi

This work was supported by the DARPA GARD Program HR001119S0026-GARD-FP-052.

\ifCLASSOPTIONcaptionsoff
  \newpage
\fi


\bibliographystyle{IEEEtran}
\bibliography{mycite}

\begin{thebibliography}{10}
\providecommand{\url}[1]{#1}
\csname url@samestyle\endcsname
\providecommand{\newblock}{\relax}
\providecommand{\bibinfo}[2]{#2}
\providecommand{\BIBentrySTDinterwordspacing}{\spaceskip=0pt\relax}
\providecommand{\BIBentryALTinterwordstretchfactor}{4}
\providecommand{\BIBentryALTinterwordspacing}{\spaceskip=\fontdimen2\font plus
\BIBentryALTinterwordstretchfactor\fontdimen3\font minus
  \fontdimen4\font\relax}
\providecommand{\BIBforeignlanguage}[2]{{%
\expandafter\ifx\csname l@#1\endcsname\relax
\typeout{** WARNING: IEEEtran.bst: No hyphenation pattern has been}%
\typeout{** loaded for the language `#1'. Using the pattern for}%
\typeout{** the default language instead.}%
\else
\language=\csname l@#1\endcsname
\fi
#2}}
\providecommand{\BIBdecl}{\relax}
\BIBdecl

\bibitem{salehi2021arae}
M.~Salehi, A.~Arya, B.~Pajoum, M.~Otoofi, A.~Shaeiri, M.~H. Rohban, and H.~R.
  Rabiee, ``Arae: Adversarially robust training of autoencoders improves
  novelty detection,'' in \emph{Neural Networks}, 2021.

\bibitem{vincent2008extracting}
P.~Vincent, H.~Larochelle, Y.~Bengio, and P.-A. Manzagol, ``Extracting and
  composing robust features with denoising autoencoders,'' in
  \emph{International Conference on Machine learning}, 2008.

\bibitem{kingma2013auto}
D.~P. Kingma and M.~Welling, ``Auto-encoding variational bayes,'' in
  \emph{International Conference on Learning Representations}, 2014.

\bibitem{makhzani2015adversarial}
A.~Makhzani, J.~Shlens, N.~Jaitly, I.~Goodfellow, and B.~Frey, ``Adversarial
  autoencoders,'' in \emph{International Conference on Learning Representations
  Workshop}, 2016.

\bibitem{pidhorskyi2018generative}
S.~Pidhorskyi, R.~Almohsen, and G.~Doretto, ``Generative probabilistic novelty
  detection with adversarial autoencoders,'' in \emph{Conference on Neural
  Information Processing Systems}, 2018.

\bibitem{goodfellow2014generative}
I.~Goodfellow, J.~Pouget-Abadie, M.~Mirza, B.~Xu, D.~Warde-Farley, S.~Ozair,
  A.~Courville, and Y.~Bengio, ``Generative adversarial nets,'' in
  \emph{Conference on Neural Information Processing Systems}, 2014.

\bibitem{perera2019ocgan}
P.~Perera, R.~Nallapati, and B.~Xiang, ``Ocgan: One-class novelty detection
  using gans with constrained latent representations,'' in \emph{IEEE
  Conference on Computer Vision and Pattern Recognition}, 2019.

\bibitem{sabokrou2018adversarially}
M.~Sabokrou, M.~Khalooei, M.~Fathy, and E.~Adeli, ``Adversarially learned
  one-class classifier for novelty detection,'' in \emph{IEEE Conference on
  Computer Vision and Pattern Recognition}, 2018.

\bibitem{salehi2020puzzle}
M.~Salehi, A.~Eftekhar, N.~Sadjadi, M.~H. Rohban, and H.~R. Rabiee,
  ``Puzzle-ae: Novelty detection in images through solving puzzles,''
  \emph{arXiv preprint arXiv:2008.12959}, 2020.

\bibitem{zhangp}
Z.~Zhang, S.~Chen, and L.~Sun, ``P-kdgan: Progressive knowledge distillation
  with gans for one-class novelty detection,'' in \emph{International Joint
  Conference on Artificial Intelligence}, 2020.

\bibitem{goodfellow2015explaining}
I.~J. Goodfellow, J.~Shlens, and C.~Szegedy, ``Explaining and harnessing
  adversarial examples,'' in \emph{International Conference on Learning
  Representations}, 2015.

\bibitem{Szegedy2014Intriguing}
C.~Szegedy, W.~Zaremba, I.~Sutskever, J.~Bruna, D.~Erhan, I.~Goodfellow, and
  R.~Fergus, ``Intriguing properties of neural networks,'' in
  \emph{International Conference on Learning Representations}, 2014.

\bibitem{guo2017countering}
C.~Guo, M.~Rana, M.~Cisse, and L.~Van Der~Maaten, ``Countering adversarial
  images using input transformations,'' in \emph{International Conference on
  Learning Representations}, 2018.

\bibitem{raff2019barrage}
E.~Raff, J.~Sylvester, S.~Forsyth, and M.~McLean, ``Barrage of random
  transforms for adversarially robust defense,'' in \emph{IEEE Conference on
  Computer Vision and Pattern Recognition}, 2019.

\bibitem{Xie_2019_CVPR}
C.~Xie, Y.~Wu, L.~van~der Maaten, A.~Yuille, and K.~He, ``Feature denoising for
  improving adversarial robustness,'' in \emph{IEEE Conference on Computer
  Vision and Pattern Recognition}, 2019.

\bibitem{xu2017feature}
W.~Xu, D.~Evans, and Y.~Qi, ``Feature squeezing: Detecting adversarial examples
  in deep neural networks,'' in \emph{Network and Distributed Systems Security
  Symposium}, 2018.

\bibitem{lo2021defending}
S.-Y. Lo and V.~M. Patel, ``Defending against multiple and unforeseen
  adversarial videos,'' \emph{IEEE Transactions on Image Processing}, 2021.

\bibitem{wei2019sparse}
X.~Wei, J.~Zhu, S.~Yuan, and H.~Su, ``Sparse adversarial perturbations for
  videos,'' in \emph{AAAI Conference on Artificial Intelligence}, 2019.

\bibitem{ranjan2019attacking}
A.~Ranjan, J.~Janai, A.~Geiger, and M.~J. Black, ``Attacking optical flow,'' in
  \emph{IEEE International Conference on Computer Vision}, 2019.

\bibitem{shao2020open}
R.~Shao, P.~Perera, P.~C. Yuen, and V.~M. Patel, ``Open-set adversarial
  defense,'' in \emph{European Conference on Computer Vision}, 2020.

\bibitem{hendrycks2019selfsupervised}
D.~Hendrycks, M.~Mazeika, S.~Kadavath, and D.~Song, ``Using self-supervised
  learning can improve model robustness and uncertainty,'' in \emph{Conference
  on Neural Information Processing Systems}, 2019.

\bibitem{shi2021online}
C.~Shi, C.~Holtz, and G.~Mishne, ``Online adversarial purification based on
  self-supervised learning,'' in \emph{International Conference on Learning
  Representations}, 2021.

\bibitem{xie2020smooth}
C.~Xie, M.~Tan, B.~Gong, A.~Yuille, and Q.~V. Le, ``Smooth adversarial
  training,'' \emph{arXiv preprint arXiv:2006.14536}, 2020.

\bibitem{park2020learning}
H.~Park, J.~Noh, and B.~Ham, ``Learning memory-guided normality for anomaly
  detection,'' in \emph{IEEE Conference on Computer Vision and Pattern
  Recognition}, 2020.

\bibitem{ross2008incremental}
D.~A. Ross, J.~Lim, R.-S. Lin, and M.-H. Yang, ``Incremental learning for
  robust visual tracking,'' in \emph{International Journal of Computer Vision},
  2008.

\bibitem{van2017neural}
A.~Van Den~Oord, O.~Vinyals \emph{et~al.}, ``Neural discrete representation
  learning,'' in \emph{Conference on Neural Information Processing Systems},
  2017.

\bibitem{madry2018towards}
A.~Madry, A.~Makelov, L.~Schmidt, D.~Tsipras, and A.~Vladu, ``Towards deep
  learning models resistant to adversarial attacks,'' in \emph{International
  Conference on Learning Representations}, 2018.

\bibitem{tsipras2018robustness}
D.~Tsipras, S.~Santurkar, L.~Engstrom, A.~Turner, and A.~Madry, ``Robustness
  may be at odds with accuracy,'' in \emph{International Conference on Learning
  Representations}, 2019.

\bibitem{xie2020adversarial}
C.~Xie, M.~Tan, B.~Gong, J.~Wang, A.~Yuille, and Q.~V. Le, ``Adversarial
  examples improve image recognition,'' in \emph{IEEE Conference on Computer
  Vision and Pattern Recognition}, 2020.

\bibitem{scholkopf1999support}
B.~Sch{\"o}lkopf, R.~C. Williamson, A.~Smola, J.~Shawe-Taylor, and J.~Platt,
  ``Support vector method for novelty detection,'' in \emph{Conference on
  Neural Information Processing Systems}, 1999.

\bibitem{tax2004support}
D.~M.~J. Tax and R.~P.~W. Duin, ``Support vector data description,'' in
  \emph{Machine Learning}, 2004.

\bibitem{gong2019memorizing}
D.~Gong, L.~Liu, V.~Le, B.~Saha, M.~R. Mansour, S.~Venkatesh, and A.~v.~d.
  Hengel, ``Memorizing normality to detect anomaly: Memory-augmented deep
  autoencoder for unsupervised anomaly detection,'' in \emph{IEEE International
  Conference on Computer Vision}, 2019.

\bibitem{sakurada2014anomaly}
M.~Sakurada and T.~Yairi, ``Anomaly detection using autoencoders with nonlinear
  dimensionality reduction,'' in \emph{ACM Conference on Machine Learning for
  Sensory Data Analysis Workshop}, 2014.

\bibitem{xia2015learning}
Y.~Xia, X.~Cao, F.~Wen, G.~Hua, and J.~Sun, ``Learning discriminative
  reconstructions for unsupervised outlier removal,'' in \emph{IEEE
  International Conference on Computer Vision}, 2015.

\bibitem{zhou2017anomaly}
C.~Zhou and R.~C. Paffenroth, ``Anomaly detection with robust deep
  autoencoders,'' in \emph{ACM International Conference on Knowledge Discovery
  and Data Mining}, 2017.

\bibitem{dong2018boosting}
Y.~Dong, F.~Liao, T.~Pang, H.~Su, J.~Zhu, X.~Hu, and J.~Li, ``Boosting
  adversarial attacks with momentum,'' in \emph{IEEE Conference on Computer
  Vision and Pattern Recognition}, 2018.

\bibitem{lo2021multav}
S.-Y. Lo and V.~M. Patel, ``Multav: Multiplicative adversarial videos,'' in
  \emph{IEEE International Conference on Advanced Video and Signal Based
  Surveillance}, 2021.

\bibitem{zajac2019adversarial}
M.~Zajac, K.~Zo{\l}na, N.~Rostamzadeh, and P.~O. Pinheiro, ``Adversarial
  framing for image and video classification,'' in \emph{AAAI Conference on
  Artificial Intelligence}, 2019.

\bibitem{hendrycks2016early}
D.~Hendrycks and K.~Gimpel, ``Early methods for detecting adversarial images,''
  in \emph{International Conference on Learning Representations Workshop},
  2017.

\bibitem{jere2020principal}
M.~Jere, S.~Herbig, C.~Lind, and F.~Koushanfar, ``Principal component
  properties of adversarial samples,'' in \emph{AAAI Conference on Artificial
  Intelligence}, 2020.

\bibitem{li2017adversarial}
X.~Li and F.~Li, ``Adversarial examples detection in deep networks with
  convolutional filter statistics,'' in \emph{IEEE International Conference on
  Computer Vision}, 2017.

\bibitem{bhagoji2017dimensionality}
A.~N. Bhagoji, ``Dimensionality reduction as a defense against evasion attacks
  on machine learning classifiers,'' \emph{arXiv preprint arXiv:1704.02654},
  2017.

\bibitem{carlini2017adversarial}
N.~Carlini and D.~Wagner, ``Adversarial examples are not easily detected:
  Bypassing ten detection methods,'' in \emph{ACM Workshop on Artificial
  Intelligence and Security}, 2017.

\bibitem{obfuscated}
A.~Athalye, N.~Carlini, and D.~Wagner, ``Obfuscated gradients give a false
  sense of security: Circumventing defenses to adversarial examples,'' in
  \emph{International Conference on Machine Learning}, 2018.

\bibitem{gidaris2018unsupervised}
S.~Gidaris, P.~Singh, and N.~Komodakis, ``Unsupervised representation learning
  by predicting image rotations,'' in \emph{International Conference on
  Learning Representations}, 2018.

\bibitem{goodge2020robustness}
A.~Goodge, B.~Hooi, S.~K. Ng, and W.~S. Ng, ``Robustness of autoencoders for
  anomaly detection under adversarial impact,'' in \emph{International Joint
  Conference on Artificial Intelligence}, 2020.

\bibitem{papernot2017practical}
N.~Papernot, P.~McDaniel, I.~Goodfellow, S.~Jha, Z.~B. Celik, and A.~Swami,
  ``Practical black-box attacks against machine learning,'' in \emph{ACM Asia
  Conference on Computer and Communications Security}, 2017.

\bibitem{lecun2010mnist}
Y.~LeCun and C.~Cortes, ``{MNIST} handwritten digit database,'' 2010.

\bibitem{xiao2017fashion}
H.~Xiao, K.~Rasul, and R.~Vollgraf, ``Fashion-mnist: a novel image dataset for
  benchmarking machine learning algorithms,'' \emph{arXiv preprint
  arXiv:1708.07747}, 2017.

\bibitem{krizhevsky2009learning}
A.~Krizhevsky \emph{et~al.}, ``Learning multiple layers of features from tiny
  images,'' 2009.

\bibitem{bergmann2019mvtec}
P.~Bergmann, M.~Fauser, D.~Sattlegger, and C.~Steger, ``Mvtec ad--a
  comprehensive real-world dataset for unsupervised anomaly detection,'' in
  \emph{IEEE conference on computer vision and pattern recognition}, 2019.

\bibitem{liu2018future}
W.~Liu, W.~Luo, D.~Lian, and S.~Gao, ``Future frame prediction for anomaly
  detection--a new baseline,'' in \emph{IEEE conference on computer vision and
  pattern recognition}, 2018.

\bibitem{mahadevan2010anomaly}
V.~Mahadevan, W.~Li, V.~Bhalodia, and N.~Vasconcelos, ``Anomaly detection in
  crowded scenes,'' in \emph{IEEE conference on computer vision and pattern
  recognition}, 2010.

\bibitem{buades2005non}
A.~Buades, B.~Coll, and J.-M. Morel, ``A non-local algorithm for image
  denoising,'' in \emph{IEEE Conference on Computer Vision and Pattern
  Recognition}, 2005.

\bibitem{hendrycks2016gaussian}
D.~Hendrycks and K.~Gimpel, ``Gaussian error linear units (gelus),''
  \emph{arXiv preprint arXiv:1606.08415}, 2016.

\bibitem{ioffe2015batch}
S.~Ioffe and C.~Szegedy, ``Batch normalization: Accelerating deep network
  training by reducing internal covariate shift,'' in \emph{International
  Conference on Machine Learning}, 2015.

\bibitem{tramer2020adaptive}
F.~Tramèr, N.~Carlini, W.~Brendel, and A.~Madry, ``On adaptive attacks to
  adversarial example defenses,'' in \emph{Conference on Neural Information
  Processing Systems}, 2020.

\bibitem{paszke2019pytorch}
A.~Paszke, S.~Gross, F.~Massa, A.~Lerer, J.~Bradbury, G.~Chanan, T.~Killeen,
  Z.~Lin, N.~Gimelshein, L.~Antiga \emph{et~al.}, ``Pytorch: An imperative
  style, high-performance deep learning library,'' in \emph{Conference on
  neural information processing systems}, 2019.

\bibitem{kingma2014adam}
D.~P. Kingma and J.~Ba, ``Adam: A method for stochastic optimization,'' in
  \emph{International Conference on Learning Representations}, 2015.

\bibitem{tramer2018ensemble}
F.~Tramèr, A.~Kurakin, N.~Papernot, I.~Goodfellow, D.~Boneh, and P.~McDaniel,
  ``Ensemble adversarial training: Attacks and defenses,'' in
  \emph{International Conference on Learning Representations}, 2018.

\bibitem{he2016deep}
K.~He, X.~Zhang, S.~Ren, and J.~Sun, ``Deep residual learning for image
  recognition,'' in \emph{IEEE conference on computer vision and pattern
  recognition}, 2016.

\bibitem{oza2020utilizing}
P.~Oza and V.~M. Patel, ``Utilizing patch-level category activation patterns
  for multiple class novelty detection,'' in \emph{European Conference on
  Computer Vision}, 2020.

\bibitem{perera2019deep}
P.~Perera and V.~M. Patel, ``Deep transfer learning for multiple class novelty
  detection,'' in \emph{IEEE Conference on Computer Vision and Pattern
  Recognition}, 2019.

\end{thebibliography}

\begin{IEEEbiography}[{\includegraphics[width=1in,height=1.25in,clip,keepaspectratio]{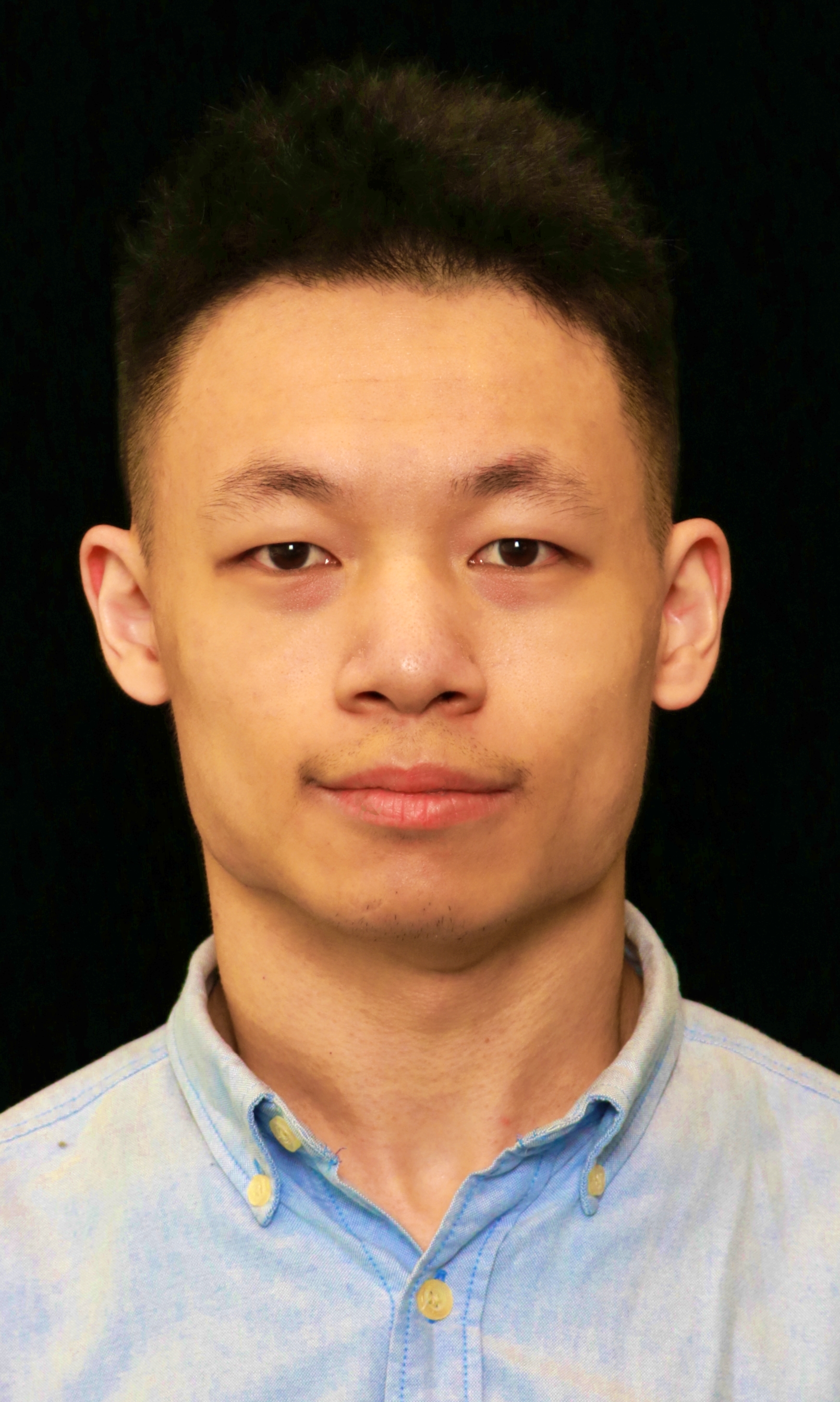}}]{Shao-Yuan Lo}
	(Student Member, IEEE) is a Ph.D. student in the Department of Electrical and Computer Engineering at Johns Hopkins University. He received his B.S. and M.S. degrees from National Chiao Tung University, Taiwan, in 2017 and 2019, respectively. His research interests include adversarial machine learning, domain adaptation and semantic segmentation. He received the Best Paper Award at ACM Multimedia Asia 2019 and the 2019 IPPR Best Master Thesis Award.
\end{IEEEbiography}

\begin{IEEEbiography}[{\includegraphics[width=1in,height=1.25in,clip,keepaspectratio]{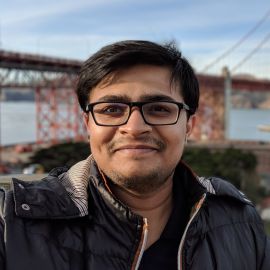}}]{Poojan Oza}
	(Student Member, IEEE) is a Ph.D. student in the Department of Electrical and Computer Engineering at Johns Hopkins University. He graduated from IIIT-Delhi with a Master’s degree in Electronics and Computer Engineering. His research interests include deep learning based one-class methods, anomaly/novelty detection, open-set recognition, domain adaptation and object detection.
\end{IEEEbiography}

\begin{IEEEbiography}[{\includegraphics[width=1in,height=1.25in,clip,keepaspectratio]{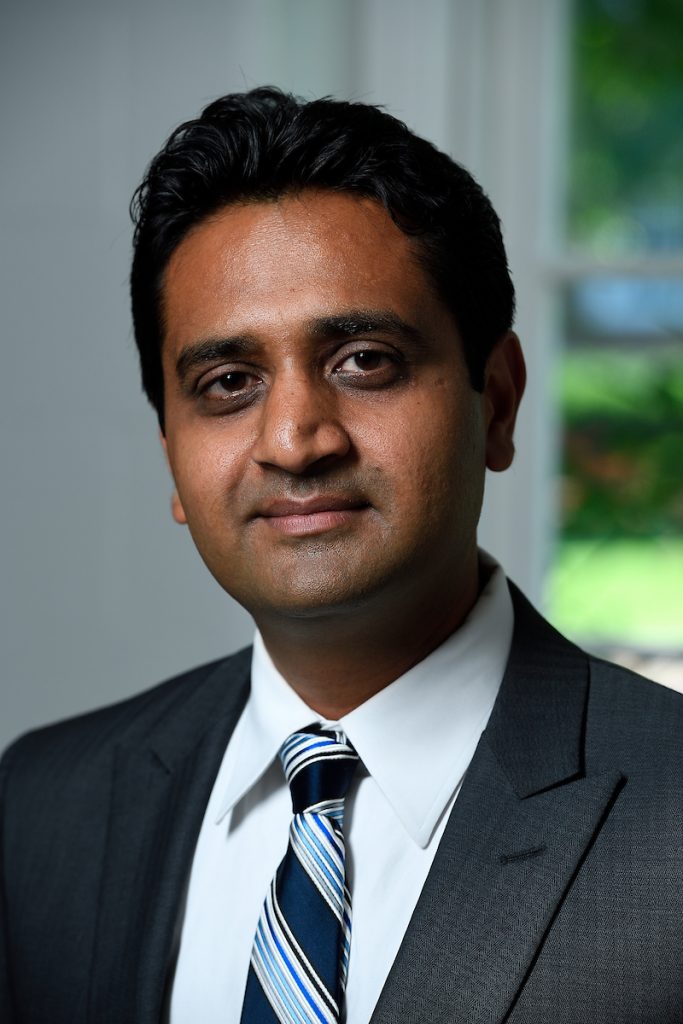}}]{Vishal M. Patel}
	(Senior Member, IEEE) is an Associate Professor in the Department of Electrical and Computer Engineering (ECE) at Johns Hopkins University. Prior to joining Hopkins, he was an A. Walter Tyson Assistant Professor in the Department of ECE at Rutgers University and a member of the research faculty at the University of Maryland Institute for Advanced Computer Studies (UMIACS). He completed his Ph.D. in Electrical Engineering from the University of Maryland, College Park, MD, in 2010. He has received a number of awards including the 2021 IEEE Signal Processing Society (SPS) Pierre-Simon Laplace Early Career Technical Achievement Award, the 2021 NSF CAREER Award, the 2021 NSF CAREER Award, the 2021 IAPR Young Biometrics Investigator Award (YBIA), the 2016 ONR Young Investigator Award, the 2016 Jimmy Lin Award for Invention, A. Walter Tyson Assistant Professorship Award, Best Paper Awards at IEEE AVSS 2017 \& 2019, IEEE BTAS 2015, IAPR ICB 2018, IEEE ICIP 2021, and two Best Student Paper Awards at IAPR ICPR 2018. He is an Associate Editor of the IEEE Transactions on Pattern Analysis and Machine Intelligence, Pattern Recognition Journal, and serves on the Machine Learning for Signal Processing (MLSP) Committee of the IEEE Signal Processing Society. He serves as the vice president of conferences for the IEEE Biometrics Council.
\end{IEEEbiography}


%------------------------------------------------------------------------
\clearpage
\setcounter{section}{0}
\renewcommand\thesection{A\arabic{section}}

%{\huge \noindent \textbf{Supplementary Material}}

\section{Basic sanity checks to evaluation}
To further verify that the proposed PrincipaLS's robustness is not due to obfuscated gradients, we report our results on the basic sanity checks introduced in Athalye et al. \cite{obfuscated}.
\begin{itemize}
	\item Table~\ref{table:main_results_1} shows that iterative attacks (PGD \cite{madry2018towards} and MI-FGSM \cite{dong2018boosting}) are stronger than one-step attacks (FGSM \cite{goodfellow2015explaining}).
	\item Table~\ref{table:main_results_1} shows that white-box attacks are stronger than black-box attacks (by MI-FGSM).
	\item Unbounded attacks reach 100\% attack success rate (AUROC drops to 0.000) on all the five datasets.
	\item Fig.~\ref{fig:epsilon} shows that increasing distortion bound increases attack success (decreases AUROC). 
\end{itemize}

\section{More on attack formulation}  \label{sec:more_attacks}
In Sec.~\ref{sec3}, we take PGD \cite{madry2018towards} as an example to illustrate the proposed attacking method against novelty detection models. Here, we elaborate on the formulation of the other attacks we used in this paper, including MI-FGSM \cite{dong2018boosting}, AF \cite{zajac2019adversarial} and MultAdv \cite{lo2021multav}.

Consider an AE-based target model with an encoder $Enc$ and a decoder $Dec$, and an input image $\mathbf{X}$ with the ground-truth label $y \in \{-1, 1\}$, where ``$1$" denotes the known class and ``$-1$" denotes the novel classes. MI-FGSM generates the adversarial example $\mathbf{X}_{adv}$ as follows:
% MI-FGSM 1
\begin{equation}
\label{mi_sgsm_1}
\mathbf{g}^{t+1} = \mu \cdot \mathbf{g}^t + \frac{\bigtriangledown_{\mathbf{X}^t} \mathcal{L}(\hat{\mathbf{X}}^t, \mathbf{X}^t, y)}{\parallel \bigtriangledown_{\mathbf{X}^t} \mathcal{L}(\hat{\mathbf{X}}^t, \mathbf{X}^t, y) \parallel_1},
\end{equation}
where $\mathbf{g}^t$ gathers the gradients of the first $t$ iterations with a decay factor $\mu$. $\mathcal{L}$ corresponds to the MSE loss defined in Eq.~\eqref{mse_loss}. Then,
% MI-FGSM 2
\begin{equation}
\label{mi_sgsm_2}
\mathbf{X}^{t+1} = Proj^{L_\infty}_{\mathbf{X}, \ \epsilon} \big\{ \mathbf{X}^{t} + \alpha \cdot sign(\mathbf{g}^{t+1}) \big\},
\end{equation}
where, $\hat{\mathbf{X}}^t = Dec(Enc(\mathbf{X}^t))$, $\alpha>0$ denotes a step size, and $t \in [0,t_{max}-1]$ is the number of attack iterations, $\mathbf{X} = \mathbf{X}^0$ and $\mathbf{X}_{adv} = \mathbf{X}^{t_{max}}$. $Proj^{L_\infty}_{\mathbf{X},\epsilon}\{\cdot\}$ projects its element into an $L_\infty$-norm bound with perturbation size $\epsilon$ such that $\parallel \mathbf{X}^{t+1} - \mathbf{X} \parallel_{\infty} \leq \epsilon$.

AF adds adversarial perturbations on the border of an images, while the remaining pixels are kept unchanged. We generate the AF example as follows:
% AF attack
\begin{equation}
\label{af_attack}
\mathbf{X}^{t+1} = Proj^{\ell_\infty}_{\mathbf{X},\epsilon} \big\{ \mathbf{X}^{t} + \mathbf{m} \cdot \alpha \cdot sign(\bigtriangledown_{\mathbf{X}^t} \mathcal{L}(\hat{\mathbf{X}}^t, \mathbf{X}^t, y)) \big\},
\end{equation}
where $\mathbf{m} \in \{0, 1\}$ is the AF mask. Let $p$ be a pixel index of $\mathbf{m}$. If $p$ is on the border of $\mathbf{m}$ within a framing width $w_{AF}$, $m_p = 1$; otherwise, $m_p = 0$.

MultAdv produces adversarial examples via the multiplicative operation, formulated as follows:
% mult ratio
\begin{equation}
\label{mult_ratio}
\mathbf{x}^{t+1} = Proj^{RB-\ell_\infty}_{\mathbf{X},\epsilon_m} \big\{ \mathbf{X}^t \odot \alpha_m^{sign(\bigtriangledown_{\mathbf{X}^t} \mathcal{L}(\hat{\mathbf{X}}^t, \mathbf{X}^t, y))} \big\},
\end{equation}
where $\alpha_m$ is the multiplicative step size, $Proj^{RB-\ell_\infty}_{\mathbf{x},\epsilon_m}\{\cdot\}$ performs projection with ratio bound $\epsilon_m$ such that $max(\frac{\mathbf{X}^{t+1}}{\mathbf{X}},\frac{\mathbf{X}}{\mathbf{X}^{t+1}}) \leq \epsilon_m$. Eq.~\eqref{mse_loss} is used as the loss objective $\mathcal{L}$ for AF and MultAdv as well to suit the novelty detection problem setup.

\section{Evaluation with FPR at 95\% TPR}
In addition to the AUROC metric, in Table~\ref{table:fpr95tpr}, we also provide the mean of FPR at 95\% TPR comparison for different defenses on the MNIST dataset \cite{lecun2010mnist}. We observe a similar trend as that of mAUROC (see Table~\ref{table:main_results_1}). The proposed PrincipaLS outperforms all the other defense approaches.

% FPR at 95\% TPR
\setlength{\tabcolsep}{10pt}
\begin{table}[!t]
	\begin{center}
		\caption{The mean of FPR at 95\% TPR under PGD attack.}
		\label{table:fpr95tpr}
		\begin{tabular}{r | c | c}
			\hline \noalign{\smallskip} \noalign{\smallskip}
			Defense & Clean & PGD \\
			\hline \noalign{\smallskip} \noalign{\smallskip}
			PGD-AT \cite{madry2018towards} & 0.229 & 0.912 \\
			FD \cite{Xie_2019_CVPR} & 0.243 & 0.914 \\
			SAT \cite{xie2020smooth} & 0.360 & 0.916 \\
			RotNet-AT \cite{hendrycks2019selfsupervised} & 0.252 & 0.909 \\
			PrincipaLS (ours) & \textbf{0.170} & \textbf{0.803} \\
			\noalign{\smallskip} \hline
		\end{tabular}
	\end{center}
\end{table}

% VQ-VAE
\setlength{\tabcolsep}{7pt}
\begin{table}
	\begin{center}
		\caption{The mAUROC of VQ-VAE and PrincipaLS under PGD attack. ``*" denotes that PGD examples are generated from a neural network gradient approximator.}
		\label{table:vqvae}
		\begin{tabular}{r | ccc}
			\hline \noalign{\smallskip} \noalign{\smallskip}
			Defense & MNIST & F-MNIST & CIFAR-10 \\
			\noalign{\smallskip} \hline \noalign{\smallskip}
			VQ-VAE \cite{van2017neural}* & 0.542 & 0.588 & 0.248 \\
			PrincipaLS (ours) & 0.706 & 0.613 & 0.246 \\
			PrincipaLS (ours)* & 0.816 & 0.755 & 0.325 \\
			\noalign{\smallskip} \hline
		\end{tabular}
	\end{center}
\end{table}

% AE PCA
%\setlength{\tabcolsep}{9pt}
%\begin{table*}[t!]
%	\begin{center}
%		\caption{The mAUROC of AE-in-AE variants and PrincipaLS on MNIST under various adversarial attacks.}
%		\label{table:aepca}
%		\begin{tabular}{r | c | ccccc | c | c}
%			\hline \noalign{\smallskip} \noalign{\smallskip}
%			Defense & Clean & FGSM \cite{goodfellow2015explaining} & PGD \cite{madry2018towards} & MI-FGSM \cite{dong2018boosting} & MultAdv \cite{lo2021multav} & AF \cite{zajac2019adversarial} & Black-box \cite{papernot2017practical} & Average \\
%			\noalign{\smallskip} \hline \noalign{\smallskip}
%			AE-in-AE-ReLU & 0.935 & 0.695 & 0.535 & 0.490 & 0.570 & 0.546 & 0.778 & 0.650 \\
%			AE-in-AE-linear & 0.972 & 0.802 & 0.681 & 0.683 & 0.709 & \textbf{0.668} & 0.851 & 0.767 \\
%			PrincipaLS (ours) & \textbf{0.973} & \textbf{0.812} & \textbf{0.706} & \textbf{0.707} & \textbf{0.725} & 0.636 & \textbf{0.866} & \textbf{0.775} \\
%			\noalign{\smallskip} \hline
%		\end{tabular}
%	\end{center}
%\end{table*}

% AE on MVTec
\setlength{\tabcolsep}{1pt}
\begin{table*}[t!]
	\begin{center}
		\caption{The AE novelty detector's AUROC of each class with different defense approaches on MVTec-AD (expansion of Table~\ref{table:main_results_1}).}
		\label{table:ae_mvtec}
		\begin{tabular}{r | r | c | ccccccccccccccc | c}
			\hline \noalign{\smallskip} \noalign{\smallskip}
			Dataset & Defense & Test type & bottle & cable & capsule & carpet & grid & hazelnut & leather & metal & pill & screw & tile & toothbrush & transistor & wood & zipper & Mean \\
			\noalign{\smallskip} \hline \noalign{\smallskip}
			& No Defense & Clean & 0.870 & 0.676 & 0.777 & 0.320 & 0.875 & 0.756 & 0.656 & 0.488 & 0.663 & 0.332 & 0.694 & 0.618 & 0.660 & 0.914 & 0.700 & 0.667 \\
			& No Defense & PGD & 0.000 & 0.000 & 0.000 & 0.000 & 0.243 & 0.204 & 0.000 & 0.000 & 0.000 & 0.000 & 0.009 & 0.000 & 0.001 & 0.058 & 0.000 & 0.034 \\
			\noalign{\smallskip} \cline{2-19} \noalign{\smallskip}
			MVTec-AD & PGD-AT \cite{madry2018towards} & & 0.000 & 0.000 & 0.000 & 0.000 & 0.483 & 0.222 & 0.000 & 0.000 & 0.000 & 0.000 & 0.009 & 0.000 & 0.000 & 0.091 & 0.000 & 0.053 \\
			& FD \cite{Xie_2019_CVPR} & PGD & 0.000 & 0.000 & 0.000 & 0.000 & 0.526 & 0.222 & 0.000 & 0.000 & 0.000 & 0.000 & 0.039 & 0.000 & 0.003 & 0.119 & 0.000 & 0.061 \\
			& PrincipaLS (ours) & & \textbf{0.235} & \textbf{0.355} & \textbf{0.195} & \textbf{0.032} & \textbf{0.578} & \textbf{0.254} & \textbf{0.080} & \textbf{0.083} & \textbf{0.037} & \textbf{0.547} & \textbf{0.092} & \textbf{0.111} & \textbf{0.314} & \textbf{0.524} & \textbf{0.213} & \textbf{0.243} \\
			\noalign{\smallskip} \hline
		\end{tabular}
	\end{center}
\end{table*}

\section{Comparison with vector quantization}
The proposed PrincipaLS learns a principal latent vector, which is adversary-free, to replace perturbed latent vectors and enhance adversarial robustness. An alternative way of learning the adversary-free latent vectors is using vector quantization. VQ-VAE \cite{van2017neural} is an AE variant that uses the vector quantization technique to improve generation ability. To the best of our knowledge, VQ-VAE has not been adopted in the context of novelty detection. In this section, we implement VQ-VAE for one-class novelty detection and evaluate its adversarial robustness. We set the number of embeddings to 4 for MNIST, 8 for F-MNIST and 256 for CIFAR-10. These numbers achieve the best robustness according to our experiments.

Because the quantization step is non-differentiable, it causes obfuscated gradients \cite{obfuscated}. Hence, we build a neural network, which consists of four fully connected layers, to learn the mapping from the latent vectors (the output of the encoder) to the quantized latent vectors (corresponded embedding vectors). Since the neural network is differentiable, we use it to approximate the gradients of the non-differentiable part to perform PGD attack \cite{madry2018towards}. For comparison, we train another neural network with the same architecture to learn the mapping from the latent space to the principal latent space of PrincipaLS.

Table~\ref{table:vqvae} reports the experimental results. Comparing PrincipaLS (PGD examples are generated from the entire differentiable network) and PrincipaLS* (PGD examples are generated from the neural network gradient approximator), we can see that the neural network still cannot perfectly approximate the gradients, so the produced attack is weaker. However, although attacked by this weaker attack, VQ-VAE achieves lower mAUROC than PrincipaLS on MNIST and F-MNIST, and much lower mAUROC than PrincipaLS* on all the datasets. This shows that PrincipaLS has better robustness than VQ-VAE.

The explanations are as follows. First, PrincipaLS's principal latent vector is learned by the incrementally-trained cascade PCA process, which is not only adversary-free but also contains important features that can properly substitute the original latent vectors. In contrast, VQ-VAE's embedding vectors are randomly initialized. Even using the training strategy in \cite{van2017neural}, the embedding vectors are still not close to the original latent vectors. Therefore, PrincipaLS's principal latent vector is a better adversary-free substitute. Second, after Vector-PCA, PrincipaLS's Vector-PCA map stores the scaling factors of the principal latent vector with spatial information, so we can perform Spatial-PCA on it to further remove the remaining adversaries. In contrast, the vector quantization map stores the indices of the embedding vectors, and we cannot do further operations on these indices. These demonstrate the advantages of the proposed PrincipaLS.

\section{Comparison with the defenses that use dimensionality reduction techniques}
A few studies employ vanilla PCA to counter adversarial attacks for the image classification problem. Hendrycks \& Gimpel \cite{hendrycks2016early} and Jere et al. \cite{jere2020principal} utilized PCA to detect adversarial examples. Li \& Li \cite{li2017adversarial} performed PCA in the feature domain and used a cascade classifier to detect adversarial examples. However, detection is inherently weaker than defense in terms of resisting adversarial attacks. Bhagoji et al. \cite{bhagoji2017dimensionality} mapped each input image into a dimensionality-reduced PCA space to defend against adversarial attacks, but this fails to resist white-box attacks \cite{carlini2017adversarial}. As discussed in Sec.~\ref{sec:introduction}, doing image classification requires a model containing sophisticated semantic information, and large manipulation such as dimensionality reduction would hurt the model capability. Hence, it is counterintuitive to use dimensionality reduction for robustifying image classification models.

In contrast, we target at a different downstream application, one-class novelty detection. As discussed in Sec.~\ref{sec:introduction}, novelty detection has a peculiar property that a novelty detector's latent space can be manipulated to a larger extent as long as it retains the known class information. This is natually suitable for using dimensionality reduction techniques to remove adversaries and maintain the model capability simultaneously. Furthermore, we propose a novel training scheme that learns the incrementally-trained cascade principal components in the latent space. The proposed defense method is fully differentiable at inference time, and it is highly robust to white-box attacks as shown in Sec.~\ref{sec:robustness}.

\section{Class-wise AUROC scores}  \label{sec:auroc}
In this section, we provide more detailed quantatitive results. From Table~\ref{table:ae_mvtec} to Table~\ref{table:arae}, they are parts of the expansion of Table~\ref{table:main_results_1} or Table~\ref{table:main_results_2}, where the AUROC scores of each class are reported.

% AE
\setlength{\tabcolsep}{6pt}
\begin{table*}[htp!]
	\begin{center}
		\caption{The AE novelty detector's AUROC of each class with different defense approaches (expansion of Table~\ref{table:main_results_2}).}
		\label{table:ae}
		\begin{tabular}{r | r | c | cccccccccc | c}
			\hline \noalign{\smallskip} \noalign{\smallskip}
			Dataset & Defense & Test type & 0 & 1 & 2 & 3 & 4 & 5 & 6 & 7 & 8 & 9 & Mean \\
			\noalign{\smallskip} \hline \noalign{\smallskip}
			& No Defense & Clean & 0.994 & 0.999 & 0.943 & 0.965 & 0.956 & 0.967 & 0.971 & 0.973 & 0.913 & 0.971 & 0.964 \\
			& No Defense & PGD & 0.000 & 0.463 & 0.000 & 0.000 & 0.020 & 0.000 & 0.001 & 0.026 & 0.000 & 0.002 & 0.051 \\
			\noalign{\smallskip} \cline{2-14} \noalign{\smallskip}
			MNIST & PGD-AT \cite{madry2018towards} & & 0.371 & 0.959 & 0.174 & 0.098 & 0.404 & 0.162 & 0.490 & 0.538 & 0.033 & 0.340 & 0.357 \\
			& FD \cite{Xie_2019_CVPR} & PGD & 0.412 & 0.956 & 0.177 & 0.117 & 0.419 & 0.146 & 0.489 & 0.548 & 0.070 & 0.322 & 0.366 \\
			& PrincipaLS (ours) & & \textbf{0.912} & \textbf{0.991} & \textbf{0.469} & \textbf{0.535} & \textbf{0.724} & \textbf{0.642} & \textbf{0.781} & \textbf{0.757} & \textbf{0.443} & \textbf{0.750} & \textbf{0.706} \\
			\noalign{\smallskip} \hline \noalign{\smallskip}
			& No Defense & Clean & 0.885 & 0.988 & 0.857 & 0.914 & 0.886 & 0.845 & 0.782 & 0.978 & 0.818 & 0.968 & 0.892 \\
			& No Defense & PGD & 0.022 & 0.202 & 0.014 & 0.088 & 0.012 & 0.071 & 0.022 & 0.374 & 0.008 & 0.068 & 0.088 \\
			\noalign{\smallskip} \cline{2-14} \noalign{\smallskip}
			F-MNIST & PGD-AT \cite{madry2018towards} & & 0.277 & 0.773 & 0.226 & 0.408 & 0.249 & 0.325 & 0.136 & 0.746 & 0.102 & 0.439 & 0.368 \\
			& FD \cite{Xie_2019_CVPR} & PGD & 0.289 & 0.790 & 0.253 & 0.412 & 0.320 & 0.328 & 0.148 & 0.770 & 0.095 & 0.387 & 0.379 \\
			& PrincipaLS (ours) & & \textbf{0.540} & \textbf{0.862} & \textbf{0.486} & \textbf{0.603} & \textbf{0.547} & \textbf{0.663} & \textbf{0.370} & \textbf{0.857} & \textbf{0.301} & \textbf{0.771} & \textbf{0.600} \\
			\noalign{\smallskip} \hline \noalign{\smallskip}
			& No Defense & Clean & 0.628 & 0.311 & 0.667 & 0.539 & 0.728 & 0.533 & 0.633 & 0.445 & 0.665 & 0.351 & 0.550 \\
			& No Defense & PGD & 0.052 & 0.003 & 0.053 & 0.022 & 0.087 & 0.020 & 0.031 & 0.013 & 0.051 & 0.006 & 0.034 \\
			\noalign{\smallskip} \cline{2-14} \noalign{\smallskip}
			CIFAR-10 & PGD-AT \cite{madry2018towards} & & 0.213 & 0.029 & 0.222 & 0.122 & 0.267 & 0.104 & 0.163 & 0.075 & 0.217 & 0.040 & 0.145 \\
			& FD \cite{Xie_2019_CVPR} & PGD & 0.211 & 0.034 & 0.225 & 0.125 & 0.269 & 0.104 & 0.171 & 0.075 & 0.212 & 0.042 & 0.147 \\
			& PrincipaLS (ours) & & \textbf{0.325} & \textbf{0.096} & \textbf{0.317} & \textbf{0.194} & \textbf{0.392} & \textbf{0.184} & \textbf{0.334} & \textbf{0.164} & \textbf{0.347} & \textbf{0.101} & \textbf{0.245} \\
			\noalign{\smallskip} \hline
		\end{tabular}
	\end{center}
\end{table*}

% VAE
\setlength{\tabcolsep}{6pt}
\begin{table*}[htp!]
	\begin{center}
		\caption{The VAE \cite{kingma2013auto} novelty detector's AUROC of each class with different defense approaches (expansion of Table~\ref{table:main_results_2}).}
		\label{table:vae}
		\begin{tabular}{r | r | c | cccccccccc | c}
			\hline \noalign{\smallskip} \noalign{\smallskip}
			Dataset & Defense & Test type & 0 & 1 & 2 & 3 & 4 & 5 & 6 & 7 & 8 & 9 & Mean \\
			\noalign{\smallskip} \hline \noalign{\smallskip}
			& No Defense & Clean & 0.998 & 0.999 & 0.971 & 0.976 & 0.976 & 0.978 & 0.993 & 0.984 & 0.934 & 0.985 & 0.979 \\
			& No Defense & PGD & 0.003 & 0.673 & 0.001 & 0.002 & 0.023 & 0.002 & 0.028 & 0.128 & 0.000 & 0.015 & 0.087 \\
			\noalign{\smallskip} \cline{2-14} \noalign{\smallskip}
			MNIST & PGD-AT \cite{madry2018towards} & & 0.679 & 0.979 & 0.382 & 0.258 & 0.598 & 0.330 & 0.687 & 0.654 & 0.159 & 0.481 & 0.521 \\
			& FD \cite{Xie_2019_CVPR} & PGD & 0.705 & 0.984 & 0.380 & 0.282 & 0.577 & 0.311 & 0.677 & 0.662 & 0.178 & 0.498 & 0.525 \\
			& PrincipaLS (ours) & & \textbf{0.921} & \textbf{0.992} & \textbf{0.622} & \textbf{0.539} & \textbf{0.760} & \textbf{0.622} & \textbf{0.845} & \textbf{0.811} & \textbf{0.501} & \textbf{0.780} & \textbf{0.739} \\
			\noalign{\smallskip} \hline \noalign{\smallskip}
			& No Defense & Clean & 0.908 & 0.990 & 0.882 & 0.934 & 0.901 & 0.888 & 0.813 & 0.983 & 0.859 & 0.979 & 0.914 \\
			& No Defense & PGD & 0.067 & 0.435 & 0.059 & 0.208 & 0.083 & 0.282 & 0.049 & 0.680 & 0.028 & 0.337 & 0.223 \\
			\noalign{\smallskip} \cline{2-14} \noalign{\smallskip}
			F-MNIST & PGD-AT \cite{madry2018towards} & & 0.460 & 0.839 & 0.392 & 0.573 & 0.475 & 0.581 & 0.305 & 0.834 & 0.193 & 0.732 & 0.538 \\
			& FD \cite{Xie_2019_CVPR} & PGD & 0.445 & 0.826 & 0.407 & 0.562 & 0.452 & 0.597 & 0.298 & 0.834 & 0.194 & 0.718 & 0.533 \\
			& PrincipaLS (ours) & & \textbf{0.536} & \textbf{0.845} & \textbf{0.477} & \textbf{0.637} & \textbf{0.524} & \textbf{0.698} & \textbf{0.381} & \textbf{0.849} & \textbf{0.315} & \textbf{0.782} & \textbf{0.604} \\
			\noalign{\smallskip} \hline \noalign{\smallskip}
			& No Defense & Clean & 0.635 & 0.336 & 0.661 & 0.522 & 0.725 & 0.512 & 0.639 & 0.463 & 0.672 & 0.354 & 0.552 \\
			& No Defense & PGD & 0.100 & 0.016 & 0.133 & 0.055 & 0.124 & 0.049 & 0.079 & 0.032 & 0.123 & 0.018 & 0.073 \\
			\noalign{\smallskip} \cline{2-14} \noalign{\smallskip}
			CIFAR-10 & PGD-AT \cite{madry2018towards} & & 0.255 & 0.050 & 0.256 & 0.146 & 0.328 & 0.123 & 0.219 & 0.092 & 0.251 & 0.052 & 0.177 \\
			& FD \cite{Xie_2019_CVPR} & PGD & 0.274 & 0.059 & 0.269 & 0.156 & 0.305 & 0.124 & 0.204 & 0.096 & 0.267 & 0.051 & 0.180 \\
			& PrincipaLS (ours) & & \textbf{0.325} & \textbf{0.097} & \textbf{0.324} & \textbf{0.198} & \textbf{0.386} & \textbf{0.197} & \textbf{0.322} & \textbf{0.171} & \textbf{0.352} & \textbf{0.101} & \textbf{0.247} \\
			\noalign{\smallskip} \hline
		\end{tabular}
	\end{center}
\end{table*}

% AAE
\setlength{\tabcolsep}{6pt}
\begin{table*}[htp!]
	\begin{center}
		\caption{The AAE \cite{makhzani2015adversarial} novelty detector's AUROC of each class with different defense approaches (expansion of Table~\ref{table:main_results_2}).}
		\label{table:aae}
		\begin{tabular}{r | r | c | cccccccccc | c}
			\hline \noalign{\smallskip} \noalign{\smallskip}
			Dataset & Defense & Test type & 0 & 1 & 2 & 3 & 4 & 5 & 6 & 7 & 8 & 9 & Mean \\
			\noalign{\smallskip} \hline \noalign{\smallskip}
			& No Defense & Clean & 0.998 & 0.999 & 0.951 & 0.974 & 0.973 & 0.962 & 0.993 & 0.976 & 0.920 & 0.983 & 0.973 \\
			& No Defense & PGD & 0.001 & 0.492 & 0.000 & 0.001 & 0.022 & 0.003 & 0.002 & 0.033 & 0.001 & 0.004 & 0.056 \\
			\noalign{\smallskip} \cline{2-14} \noalign{\smallskip}
			MNIST & PGD-AT \cite{madry2018towards} & & 0.509 & 0.965 & 0.167 & 0.185 & 0.537 & 0.208 & 0.590 & 0.569 & 0.143 & 0.401 & 0.427 \\
			& FD \cite{Xie_2019_CVPR} & PGD & 0.592 & 0.968 & 0.138 & 0.145 & 0.474 & 0.209 & 0.556 & 0.593 & 0.134 & 0.386 & 0.419 \\
			& PrincipaLS (ours) & & \textbf{0.727} & \textbf{0.985} & \textbf{0.432} & \textbf{0.410} & \textbf{0.659} & \textbf{0.419} & \textbf{0.763} & \textbf{0.723} & \textbf{0.311} & \textbf{0.649} & \textbf{0.608} \\
			\noalign{\smallskip} \hline \noalign{\smallskip}
			& No Defense & Clean & 0.908 & 0.988 & 0.875 & 0.930 & 0.900 & 0.887 & 0.819 & 0.986 & 0.852 & 0.975 & 0.912 \\
			& No Defense & PGD & 0.094 & 0.091 & 0.011 & 0.196 & 0.053 & 0.303 & 0.071 & 0.501 & 0.016 & 0.189 & 0.152 \\
			\noalign{\smallskip} \cline{2-14} \noalign{\smallskip}
			F-MNIST & PGD-AT \cite{madry2018towards} & & 0.428 & 0.761 & 0.398 & 0.551 & 0.443 & 0.546 & 0.308 & 0.817 & 0.155 & 0.711 & 0.512 \\
			& FD \cite{Xie_2019_CVPR} & PGD & 0.408 & 0.739 & 0.453 & 0.534 & 0.450 & 0.557 & 0.326 & 0.804 & 0.188 & 0.672 & 0.513 \\
			& PrincipaLS (ours) & & \textbf{0.537} & \textbf{0.795} & \textbf{0.532} & \textbf{0.609} & \textbf{0.516} & \textbf{0.660} & \textbf{0.451} & \textbf{0.832} & \textbf{0.304} & \textbf{0.748} & \textbf{0.599} \\
			\noalign{\smallskip} \hline \noalign{\smallskip}
			& No Defense & Clean & 0.634 & 0.336 & 0.661 & 0.524 & 0.732 & 0.499 & 0.662 & 0.465 & 0.671 & 0.366 & 0.555 \\
			& No Defense & PGD & 0.074 & 0.011 & 0.041 & 0.055 & 0.097 & 0.0333 & 0.080 & 0.032 & 0.071 & 0.012 & 0.051 \\
			\noalign{\smallskip} \cline{2-14} \noalign{\smallskip}
			CIFAR-10 & PGD-AT \cite{madry2018towards} & & 0.274 & 0.058 & 0.284 & 0.168 & 0.320 & 0.143 & 0.265 & 0.109 & 0.264 & 0.063 & 0.195 \\
			& FD \cite{Xie_2019_CVPR} & PGD & 0.306 & 0.065 & 0.276 & 0.161 & 0.374 & 0.139 & 0.284 & 0.123 & 0.267 & 0.068 & 0.206 \\
			& PrincipaLS (ours) & & \textbf{0.324} & \textbf{0.088} & \textbf{0.333} & \textbf{0.205} & \textbf{0.410} & \textbf{0.191} & \textbf{0.352} & \textbf{0.172} & \textbf{0.340} & \textbf{0.110} & \textbf{0.252} \\
			\noalign{\smallskip} \hline
		\end{tabular}
	\end{center}
\end{table*}

% ALOCC
\setlength{\tabcolsep}{6pt}
\begin{table*}[htp!]
	\begin{center}
		\caption{The ALOCC \cite{sabokrou2018adversarially} novelty detector's AUROC of each class with different defense approaches (expansion of Table~\ref{table:main_results_2}).}
		\label{table:alocc}
		\begin{tabular}{r | r | c | cccccccccc | c}
			\hline \noalign{\smallskip} \noalign{\smallskip}
			Dataset & Defense & Test type & 0 & 1 & 2 & 3 & 4 & 5 & 6 & 7 & 8 & 9 & Mean \\
			\noalign{\smallskip} \hline \noalign{\smallskip}
			& No Defense & Clean & 0.990 & 0.998 & 0.913 & 0.974 & 0.961 & 0.955 & 0.986 & 0.957 & 0.918 & 0.965 & 0.961 \\
			& No Defense & PGD & 0.019 & 0.742 & 0.007 & 0.018 & 0.270 & 0.008 & 0.081 & 0.194 & 0.000 & 0.071 & 0.141 \\
			\noalign{\smallskip} \cline{2-14} \noalign{\smallskip}
			MNIST & PGD-AT \cite{madry2018towards} & & 0.262 & 0.962 & 0.141 & 0.071 & 0.438 & 0.098 & 0.372 & 0.464 & 0.009 & 0.298 & 0.312 \\
			& FD \cite{Xie_2019_CVPR} & PGD & 0.306 & 0.977 & 0.127 & 0.127 & 0.352 & 0.091 & 0.402 & 0.539 & 0.023 & 0.247 & 0.319 \\
			& PrincipaLS (ours) & & \textbf{0.911} & \textbf{0.990} & \textbf{0.594} & \textbf{0.487} & \textbf{0.609} & \textbf{0.554} & \textbf{0.849} & \textbf{0.752} & \textbf{0.422} & \textbf{0.764} & \textbf{0.693} \\
			\noalign{\smallskip} \hline \noalign{\smallskip}
			& No Defense & Clean & 0.912 & 0.988 & 0.879 & 0.926 & 0.899 & 0.851 & 0.823 & 0.980 & 0.794 & 0.958 & 0.901 \\
			& No Defense & PGD & 0.094 & 0.525 & 0.062 & 0.113 & 0.102 & 0.161 & 0.052 & 0.538 & 0.023 & 0.099 & 0.177 \\
			\noalign{\smallskip} \cline{2-14} \noalign{\smallskip}
			F-MNIST & PGD-AT \cite{madry2018towards} & & 0.297 & 0.769 & 0.266 & 0.406 & 0.300 & 0.284 & 0.146 & 0.733 & 0.092 & 0.375 & 0.367 \\
			& FD \cite{Xie_2019_CVPR} & PGD & 0.316 & 0.790 & 0.230 & 0.415 & 0.316 & 0.287 & 0.149 & 0.733 & 0.087 & 0.377 & 0.370 \\
			& PrincipaLS (ours) & & \textbf{0.554} & \textbf{0.870} & \textbf{0.565} & \textbf{0.610} & \textbf{0.531} & \textbf{0.708} & \textbf{0.370} & \textbf{0.839} & \textbf{0.265} & \textbf{0.810} & \textbf{0.612} \\
			\noalign{\smallskip} \hline \noalign{\smallskip}
			& No Defense & Clean & 0.617 & 0.324 & 0.664 & 0.538 & 0.731 & 0.530 & 0.630 & 0.450 & 0.671 & 0.358 & 0.551 \\
			& No Defense & PGD & 0.061 & 0.005 & 0.068 & 0.023 & 0.081 & 0.019 & 0.035 & 0.012 & 0.063 & 0.006 & 0.037 \\
			\noalign{\smallskip} \cline{2-14} \noalign{\smallskip}
			CIFAR-10 & PGD-AT \cite{madry2018towards} & & 0.217 & 0.032 & 0.216 & 0.123 & 0.271 & 0.111 & 0.165 & 0.073 & 0.210 & 0.040 & 0.146 \\
			& FD \cite{Xie_2019_CVPR} & PGD & 0.208 & 0.035 & 0.217 & 0.126 & 0.291 & 0.106 & 0.183 & 0.072 & 0.220 & 0.045 & 0.150 \\
			& PrincipaLS (ours) & & \textbf{0.322} & \textbf{0.097} & \textbf{0.304} & \textbf{0.205} & \textbf{0.386} & \textbf{0.188} & \textbf{0.331} & \textbf{0.171} & \textbf{0.346} & \textbf{0.095} & \textbf{0.244} \\
			\noalign{\smallskip} \hline
		\end{tabular}
	\end{center}
\end{table*}

% GPND
\setlength{\tabcolsep}{6pt}
\begin{table*}[htp!]
	\begin{center}
		\caption{The GPND \cite{pidhorskyi2018generative} novelty detector's AUROC of each class with different defense approaches (expansion of Table~\ref{table:main_results_2}).}
		\label{table:gpnd}
		\begin{tabular}{r | r | c | cccccccccc | c}
			\hline \noalign{\smallskip} \noalign{\smallskip}
			Dataset & Defense & Test type & 0 & 1 & 2 & 3 & 4 & 5 & 6 & 7 & 8 & 9 & Mean \\
			\noalign{\smallskip} \hline \noalign{\smallskip}
			& No Defense & Clean & 0.998 & 0.998 & 0.911 & 0.921 & 0.894 & 0.937 & 0.984 & 0.965 & 0.914 & 0.936 & 0.946 \\
			& No Defense & PGD & 0.029 & 0.757 & 0.002 & 0.050 & 0.142 & 0.004 & 0.031 & 0.049 & 0.010 & 0.206 & 0.128 \\
			\noalign{\smallskip} \cline{2-14} \noalign{\smallskip}
			MNIST & PGD-AT \cite{madry2018towards} & & 0.706 & 0.987 & 0.507 & 0.361 & 0.670 & 0.350 & 0.716 & 0.739 & 0.207 & 0.570 & 0.582 \\
			& FD \cite{Xie_2019_CVPR} & PGD & 0.689 & 0.987 & 0.463 & 0.413 & 0.407 & 0.472 & 0.762 & 0.528 & 0.241 & 0.547 & 0.551 \\
			& PrincipaLS (ours) & & \textbf{0.880} & \textbf{0.991} & \textbf{0.557} & \textbf{0.543} & \textbf{0.766} & \textbf{0.613} & \textbf{0.821} & \textbf{0.745} & \textbf{0.443} & \textbf{0.786} & \textbf{0.741} \\
			\noalign{\smallskip} \hline \noalign{\smallskip}
			& No Defense & Clean & 0.907 & 0.981 & 0.877 & 0.911 & 0.916 & 0.900 & 0.831 & 0.983 & 0.878 & 0.963 & 0.915 \\
			& No Defense & PGD & 0.076 & 0.328 & 0.016 & 0.227 & 0.008 & 0.407 & 0.014 & 0.581 & 0.006 & 0.103 & 0.177 \\
			\noalign{\smallskip} \cline{2-14} \noalign{\smallskip}
			F-MNIST & PGD-AT \cite{madry2018towards} & & 0.487 & 0.821 & 0.409 & 0.585 & 0.452 & 0.570 & 0.340 & 0.833 & 0.198 & 0.690 & 0.539 \\
			& FD \cite{Xie_2019_CVPR} & PGD & 0.526 & 0.821 & 0.411 & 0.578 & 0.484 & 0.546 & 0.321 & 0.832 & 0.195 & 0.705 & 0.542 \\
			& PrincipaLS (ours) & & \textbf{0.586} & \textbf{0.840} & \textbf{0.530} & \textbf{0.665} & \textbf{0.547} & \textbf{0.694} & \textbf{0.424} & \textbf{0.860} & \textbf{0.331} & \textbf{0.785} & \textbf{0.626} \\
			\noalign{\smallskip} \hline \noalign{\smallskip}
			& No Defense & Clean & 0.659 & 0.344 & 0.659 & 0.520 & 0.735 & 0.507 & 0.674 & 0.464 & 0.647 & 0.358 & 0.559 \\
			& No Defense & PGD & 0.024 & 0.002 & 0.076 & 0.019 & 0.055 & 0.022 & 0.026 & 0.014 & 0.027 & 0.006 & 0.027 \\
			\noalign{\smallskip} \cline{2-14} \noalign{\smallskip}
			CIFAR-10 & PGD-AT \cite{madry2018towards} & & 0.260 & 0.052 & 0.266 & 0.139 & 0.322 & 0.127 & 0.247 & 0.107 & 0.249 & 0.055 & 0.182 \\
			& FD \cite{Xie_2019_CVPR} & PGD & 0.258 & 0.057 & 0.269 & 0.154 & 0.330 & 0.141 & 0.254 & 0.099 & 0.252 & 0.058 & 0.187 \\
			& PrincipaLS (ours) & & \textbf{0.314} & \textbf{0.095} & \textbf{0.330} & \textbf{0.195} & \textbf{0.374} & \textbf{0.183} & \textbf{0.335} & \textbf{0.167} & \textbf{0.325} & \textbf{0.098} & \textbf{0.242} \\
			\noalign{\smallskip} \hline
		\end{tabular}
	\end{center}
\end{table*}

% ARAE
\setlength{\tabcolsep}{6pt}
\begin{table*}[htp!]
	\begin{center}
		\caption{The ARAE \cite{salehi2021arae} novelty detector's AUROC of each class with different defense approaches (expansion of Table~\ref{table:main_results_2}).}
		\label{table:arae}
		\begin{tabular}{r | r | c | cccccccccc | c}
			\hline \noalign{\smallskip} \noalign{\smallskip}
			Dataset & Defense & Test type & 0 & 1 & 2 & 3 & 4 & 5 & 6 & 7 & 8 & 9 & Mean \\
			\noalign{\smallskip} \hline \noalign{\smallskip}
			& No Defense & Clean & 0.993 & 0.999 & 0.951 & 0.958 & 0.965 & 0.953 & 0.981 & 0.972 & 0.917 & 0.965 & 0.965 \\
			& No Defense & PGD & 0.022 & 0.754 & 0.004 & 0.001 & 0.013 & 0.004 & 0.050 & 0.239 & 0.000 & 0.144 & 0.133 \\
			\noalign{\smallskip} \cline{2-14} \noalign{\smallskip}
			MNIST & PGD-AT \cite{madry2018towards} & & 0.341 & 0.965 & 0.152 & 0.098 & 0.391 & 0.144 & 0.461 & 0.522 & 0.047 & 0.349 & 0.341 \\
			& FD \cite{Xie_2019_CVPR} & PGD & 0.382 & 0.950 & 0.170 & 0.106 & 0.402 & 0.108 & 0.465 & 0.545 & 0.039 & 0.336 & 0.350 \\
			& PrincipaLS (ours) & & \textbf{0.891} & \textbf{0.989} & \textbf{0.477} & \textbf{0.472} & \textbf{0.723} & \textbf{0.564} & \textbf{0.858} & \textbf{0.755} & \textbf{0.498} & \textbf{0.723} & \textbf{0.695} \\
			\noalign{\smallskip} \hline \noalign{\smallskip}
			& No Defense & Clean & 0.876 & 0.975 & 0.852 & 0.937 & 0.913 & 0.847 & 0.772 & 0.984 & 0.883 & 0.982 & 0.901 \\
			& No Defense & PGD & 0.103 & 0.487 & 0.057 & 0.415 & 0.077 & 0.272 & 0.040 & 0.803 & 0.023 & 0.345 & 0.262 \\
			\noalign{\smallskip} \cline{2-14} \noalign{\smallskip}
			F-MNIST & PGD-AT \cite{madry2018towards} & & 0.302 & 0.766 & 0.280 & 0.491 & 0.351 & 0.381 & 0.182 & 0.818 & 0.104 & 0.524 & 0.420 \\
			& FD \cite{Xie_2019_CVPR} & PGD & 0.344 & 0.784 & 0.264 & 0.510 & 0.364 & 0.404 & 0.175 & 0.798 & 0.113 & 0.525 & 0.428 \\
			& PrincipaLS (ours) & & \textbf{0.511} & \textbf{0.842} & \textbf{0.459} & \textbf{0.614} & \textbf{0.531} & \textbf{0.673} & \textbf{0.395} & \textbf{0.848} & \textbf{0.353} & \textbf{0.765} & \textbf{0.599} \\
			\noalign{\smallskip} \hline \noalign{\smallskip}
			& No Defense & Clean & 0.670 & 0.389 & 0.626 & 0.518 & 0.686 & 0.526 & 0.571 & 0.490 & 0.738 & 0.568 & 0.578 \\
			& No Defense & PGD & 0.136 & 0.014 & 0.109 & 0.070 & 0.144 & 0.088 & 0.074 & 0.060 & 0.144 & 0.028 & 0.087 \\
			\noalign{\smallskip} \cline{2-14} \noalign{\smallskip}
			CIFAR-10 & PGD-AT \cite{madry2018towards} & & 0.217 & 0.039 & 0.235 & 0.125 & 0.278 & 0.108 & 0.208 & 0.086 & 0.230 & 0.043 & 0.157 \\
			& FD \cite{Xie_2019_CVPR} & PGD & 0.223 & 0.028 & 0.240 & 0.122 & 0.275 & 0.106 & 0.189 & 0.079 & 0.216 & 0.039 & 0.152 \\
			& PrincipaLS (ours) & & \textbf{0.329} & \textbf{0.099} & \textbf{0.314} & \textbf{0.193} & \textbf{0.388} & \textbf{0.189} & \textbf{0.334} & \textbf{0.168} & \textbf{0.344} & \textbf{0.098} & \textbf{0.245} \\
			\noalign{\smallskip} \hline
		\end{tabular}
	\end{center}
\end{table*}

\end{document}